\documentclass[letterpaper]{article}
\usepackage{aaai2027}
 
\usepackage{graphicx}
 
\usepackage{natbib}
\usepackage{booktabs}
\usepackage{xcolor}
\usepackage[most]{tcolorbox}
 
\title{\textsc{Cardiologent}: Multi-Agent Clinical Decision Support for
Patient-Level Arrhythmia Assessment, Urgency, and Management}
 
\author{
    Sukju Oh\textsuperscript{\rm 1},
    Moo-Yong Rhee\textsuperscript{\rm 2},
    Jae-Sik Jang\textsuperscript{\rm 2},
    Sukkyu Sun\textsuperscript{\rm 1}\thanks{Corresponding authors.}
}
\affiliations{
    \textsuperscript{\rm 1}Department of Computer Science and Artificial
      Intelligence, Dongguk University, Seoul, 04620, Republic of Korea\\
    \textsuperscript{\rm 2}Department of Cardiology, Dongguk University Ilsan
      Hospital, Goyang, 10326, Republic of Korea\\
    dhtjrwn119@dgu.ac.kr, mooyong.rhee@gmail.com, jsjang71@gmail.com, sukkyu.sun@dgu.ac.kr
}
 
\begin{document}
\maketitle
 
\begin{abstract}
The same episode of atrial fibrillation is a minor finding in a healthy adult and
grounds for anticoagulation in an elderly patient with hypertension: identical
signal, opposite decision. Naming the rhythm is only the start; what determines a
patient's outcome is the judgement that follows---what the arrhythmia is across
the whole record, what it means for this patient, and what should be done about
it. Recent work pairing large language models with the ECG stops short of this,
reading one recording without assembling it into a patient-level finding; and
agentic systems built around it either receive the arrhythmia a device has
already detected or target a different diagnostic task, stopping before the
decision this task requires. We formulate patient-level arrhythmia decision
support as a task and present \textsc{Cardiologent}, a multi-agent system that
spans it from detection to decision. An agent for each signal---a single ECG lead
and the photoplethysmogram a wearable acquires---grounds its window reading in
measured features rather than a bare label; the readings are assembled into the
patient's rhythm profile and, with the patient's own data, reasoned against
clinical guidelines retrieved for the case, with a critic checking each
conclusion against the guideline it cites. We evaluate the clinical decision
rather than the report, across integrated diagnosis, clinical significance, and
urgency and management. \textsc{Cardiologent} scores highest on every axis, first
on every patient-level task under both cardiologists and an at-scale LLM
judge---whose agreement with the cardiologists (ICC $0.74$, $0.66$) matches theirs
with each other ($0.67$). Because each conclusion traces to a cited guideline and
is validated against expert cardiologists, it yields decisions a clinician can
audit rather than act on blindly---a step toward systems that could be used in
continuous monitoring. Code and data are available at https://github.com/sukjuoh/Cardiologent.
\end{abstract}
 
\begin{figure}[!t]
\centering
\includegraphics[width=\linewidth]{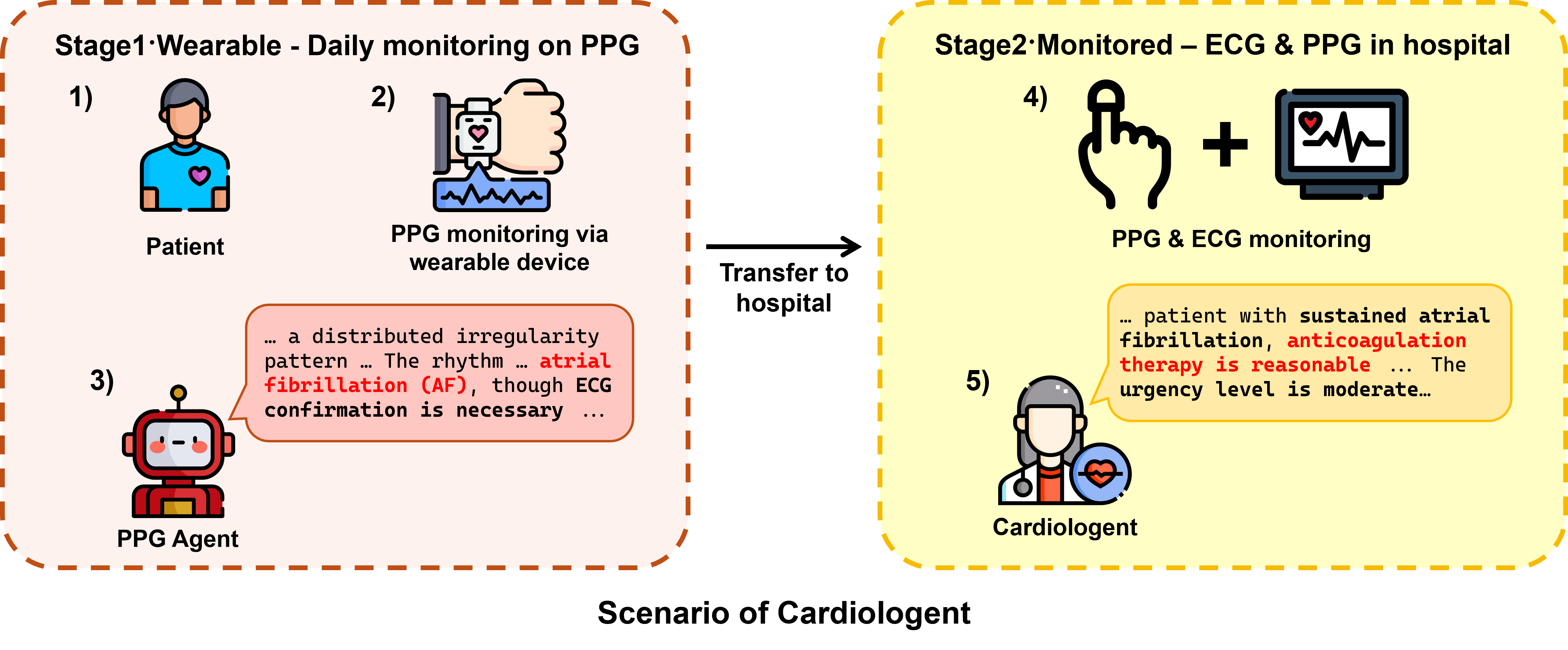}
\caption{\textbf{Deployment scenario.} \emph{Stage 1, wearable:} the pulse is
monitored on its own, and the PPG agent reads it as far as the pulse allows---an
irregular rhythm it calls atrial fibrillation, flagged as needing an ECG to
confirm. \emph{Stage 2, monitored:} the ECG is added, and \textsc{Cardiologent}
reads it together with the pulse and the patient's own data to reach the
decision---the therapy indicated, and how urgent. Excerpts are verbatim.}
\label{fig:scenario}
\end{figure}

\begin{figure*}[!t]
\centering
\includegraphics[width=\textwidth]{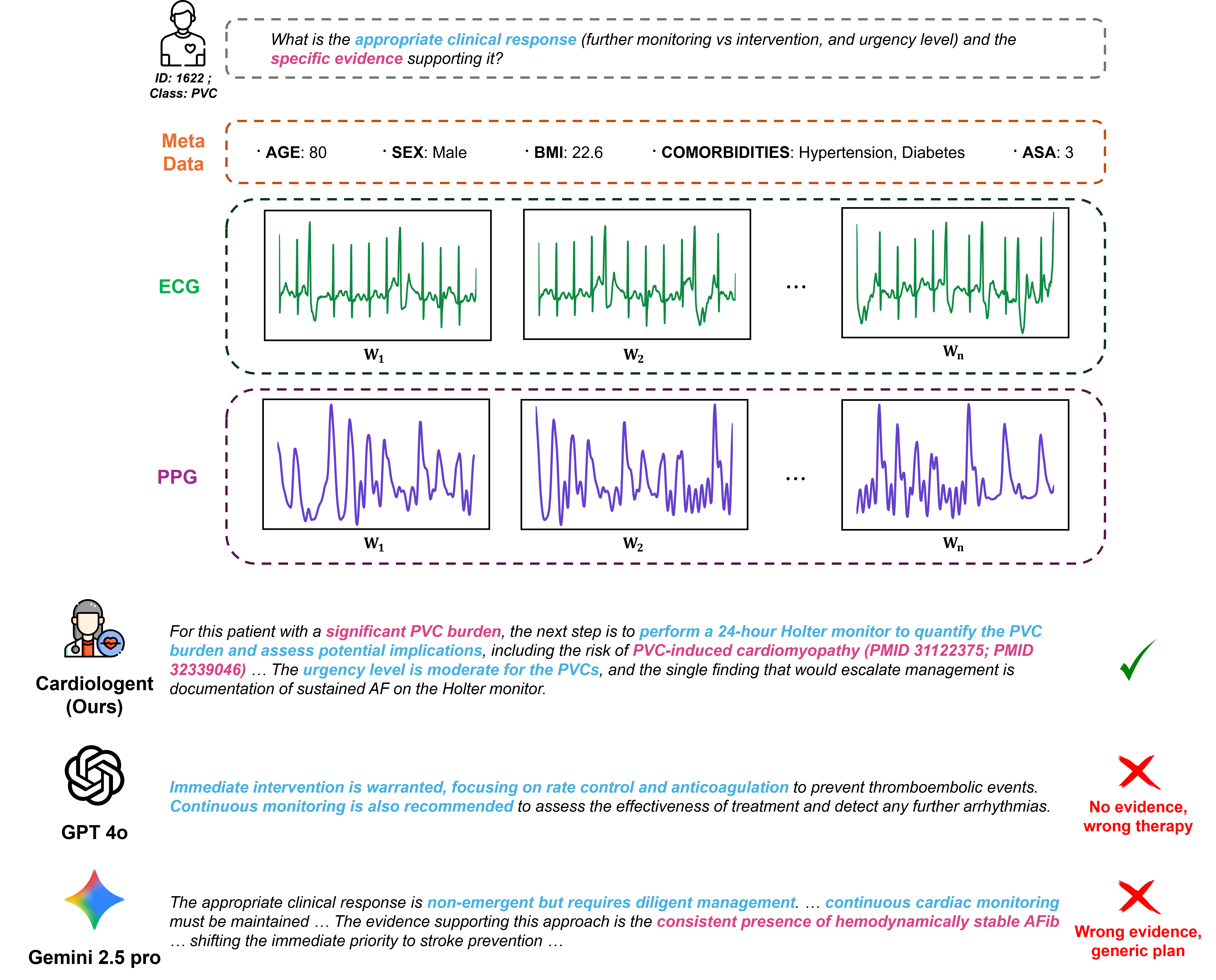}
\caption{\textbf{Patient-level arrhythmia decision support on a monitored record.} An 80-year-old man with
hypertension and diabetes, whose record holds a burden of premature ventricular
contractions (PVC). Every system gets the same input: his metadata and the windowed
lead~II ECG and PPG. Blue marks the clinical response, pink the evidence for it.
GPT-4o prescribes rate control and anticoagulation with no evidence at all, never
naming the ectopy. Gemini~2.5~Pro names a rhythm the record does not show---atrial
fibrillation---and builds on it a plan that would fit any patient.
\textsc{Cardiologent} quantifies the finding it has and cites the guideline
behind the recommendation. Excerpts are verbatim; ellipses mark omitted text.}
\label{fig:teaser}
\end{figure*}
 
\section{Introduction}
 
Cardiac arrhythmias are a major cause of cardiovascular morbidity, contributing
to stroke, heart failure, and sudden cardiac
death~\citep{fu2015arrhythmia,zeppenfeld2022esc}. Atrial fibrillation, the most
common arrhythmia, is asymptomatic in roughly 32\% of
patients~\citep{frykman2001asymptomatic} and often goes undetected until a severe event
occurs. Detecting such arrhythmias reliably requires monitoring a patient's
cardiac signals continuously.
 
Deep learning has driven much of the progress on this problem. End-to-end
networks classify arrhythmias from single- and 12-lead electrocardiogram (ECG) at
cardiologist-level accuracy~\citep{hannun2019cardiologist,Ribeiro_2020,siontis2021ai,li2025ecgfounder}, and comparable methods detect atrial fibrillation
from wearable photoplethysmogram (PPG) at population
scale~\citep{perez2019large,guo2019mobile}. These models predict a class
label but do not explain it: they give no account of which signal features
support the prediction or why competing rhythms are excluded
~\citep{wang2026position}. Many arrhythmias are separated
by narrow margins---atrial fibrillation from other irregular rhythms,
supraventricular from ventricular ectopy---and the distinction turns on
specific, localizable features of the signal. A clinician given only a label
cannot see which feature the decision rested on, and so cannot confirm or
overturn it; since these distinctions carry directly into management, an
unverifiable label is of limited clinical use.

Recent work supplies the grounds a bare label lacks, on the discovery that large
language models can be made to operate on time series at
all~\citep{zhou2023onefitsall,jin2024timellmtimeseriesforecasting,alnegheimish2024sigllm,zhou2025tsanomalies,quinlan2026chatts}. Applied to the
ECG, this produces models that pair the waveform with a language model and
explain their reading rather than only labeling it, from a textual diagnosis to
a full report grounded in the
waveform~\citep{liu2026pulse,wan2025meit,%
lan2025gem,jin2025uniecg,jin2026ecgr1,chung2026ecgagent,%
li2026anyecgchat}.
But they stop at the reading: each reports what
the signal shows, and none assembles it into a finding about the patient.

A second line of work reaches further, to the document a physician actually
reads. Zodiac assembles a patient-specific report from a monitoring study,
synthesizing findings against clinical guidelines under cardiologist
validation~\citep{zhou2024zodiac}; CardAIc-Agents ingests the raw signal and
coordinates tools over it~\citep{zhang2025cardaic}. But neither spans the task we pose. Zodiac is
handed the findings a recording device already computed and a strip already
selected to depict the significant rhythm---detection, and the choice of what to
look at, have happened before the model is invoked---and its interpretation stops
at what those findings mean, validated as a report: its accuracy against the
patient's data and its succinctness. CardAIc reads
the signal itself, but its task is disease-level diagnosis scored by
classification accuracy, not the rhythm-level reading of a monitored record
assembled into a patient-level finding; the management plans it forms along the
way are never scored.
 
Between naming a rhythm and knowing what to do about it lies most of the
clinical act, and it is not a step that follows from the waveform. The same
episode of atrial fibrillation is a minor finding in a healthy 40-year-old and
grounds for anticoagulation in an elderly woman with
hypertension~\citep{joglar2024af}: identical signal, opposite decision. What
separates them is the patient---age, sex, comorbidity, medication---and none of
it is visible in the trace. Nor is the decision made on one window. Whether an
abnormal beat is incidental or part of a sustained arrhythmia, and whether an
episode recurred or happened once, are properties of the record rather than of
any window in it, and they are what set the urgency. A patient-level judgement
therefore needs both: the record assembled, and the patient's own data alongside
it. GEM's authors observe the second directly:
without clinical context, their model overstates the severity of its
findings~\citep{lan2025gem}.

The signals such a judgement can be built on are constrained in turn. The
12-lead ECG demands electrode placement and a stationary patient, and is
ill-suited to monitoring a record over hours; what monitors record instead is a
small montage in which lead~II is the rhythm lead---its axis aligns with that of
the P~wave, so atrial activity shows itself most clearly there---alongside the
PPG, acquired continuously by clinical monitors and by every wearable on the
market. The two are not always both in hand (Figure~\ref{fig:scenario}): the
pulse runs continuously, on a ward monitor and on a wearable alike, while the ECG
is attached only when something in the pulse calls for it. A system for this
setting must therefore reach a reading from the pulse by itself, enough to say
that an ECG is needed, and a fuller one once the ECG arrives. Hence an agent per
signal rather than one reader over the pair, which is defined only when both are
present.

We present \textsc{Cardiologent}, a multi-agent system that works on a monitored
record---lead~II ECG and PPG---at two levels. At the window
level, an agent for each signal reads its own waveform through tools that measure
it---a specialist fine-tuned on that signal, and analysts for the features the
reading turns on---and grounds every claim in a measured quantity; the two
readings are then reconciled into one verdict for the window. At the patient
level, the windows are assembled into the patient's rhythm profile---which
arrhythmias appear and which dominates---and only then does the patient enter:
retrieved guidelines, a reasoning stage judging what the profile means for this
patient, what to do about it and how urgently, and a critic holding each
conclusion to the guideline it cites.
Figure~\ref{fig:teaser} shows what this is worth
on one patient: given the same record and the same metadata, GPT-4o prescribes
anticoagulation for a rhythm the record does not hold, citing nothing;
Gemini~2.5~Pro names that rhythm as its evidence and returns a plan that would read
the same for any patient; \textsc{Cardiologent} quantifies the finding actually
present and cites the guideline its recommendation comes from. Work on the PPG
has concentrated almost entirely on atrial
fibrillation~\citep{pereira2020ppg,liu2022multiclass}; our window level reads all
seven classes from the pulse alone.

In summary, the main contributions of this work are as follows:
\begin{itemize}
\item We formulate patient-level arrhythmia decision support as a task: not the
classification of a window alone, nor the interpretation of findings a device has
already computed, but the whole span between them---what arrhythmia a record
holds, what it means for this patient, and what should be done about it.
 
\item We present \textsc{Cardiologent}, the first system to span this task, as we define it,
from detection to decision. Where prior work receives its findings from a recording
device, \textsc{Cardiologent} assembles the patient's profile from its own
window-level evidence; at the window level it also runs on the PPG alone, the
signal a wearable acquires.
 
\item We evaluate the clinical decision itself, not the fluency of a report:
cardiologists score \textsc{Cardiologent} against frontier baselines on what the
system concludes and would do, the level at which such a system must ultimately be judged.
\end{itemize}

\begin{figure*}[!t]
\centering
\includegraphics[width=\textwidth]{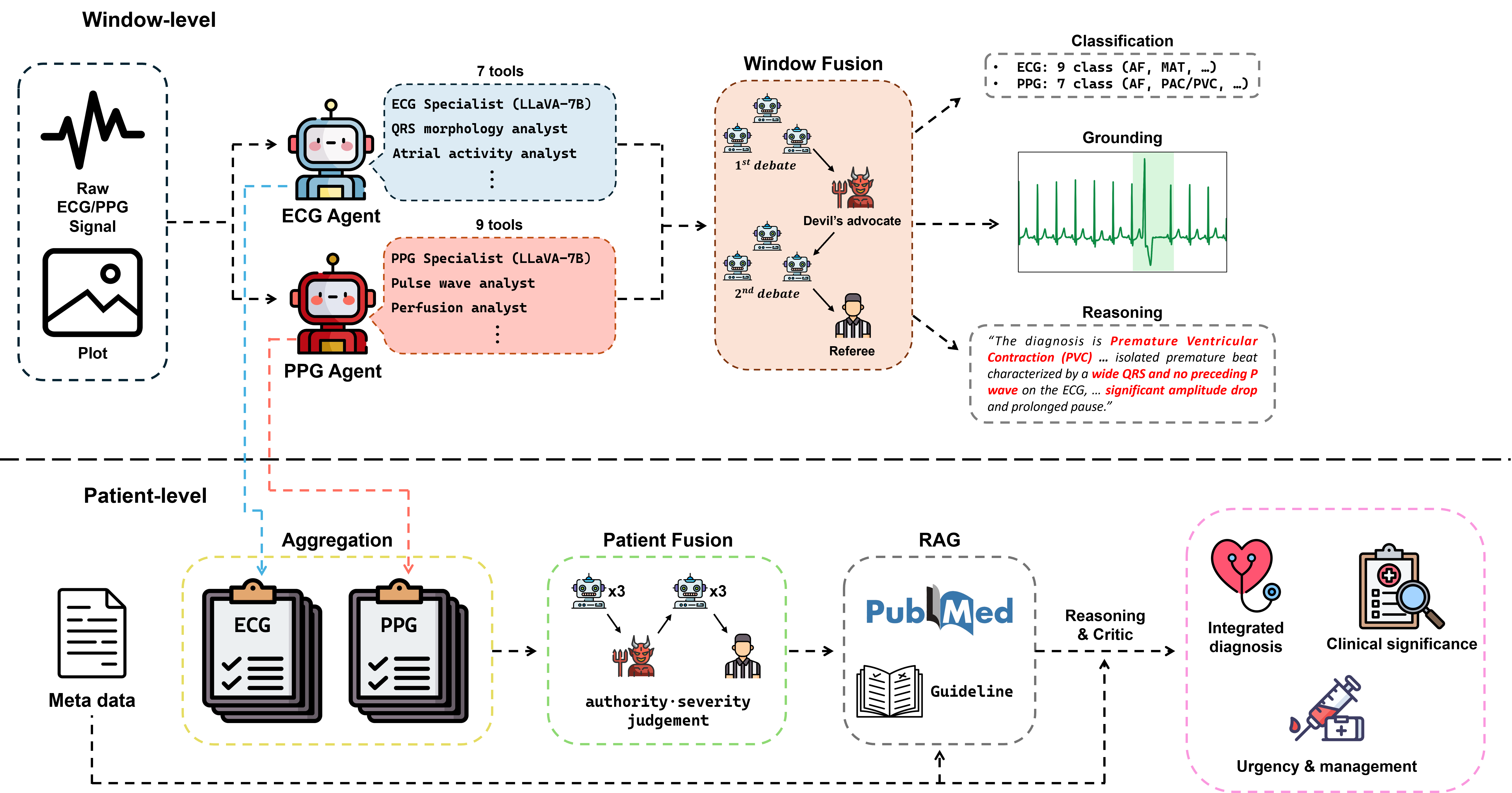}
\caption{\textbf{\textsc{Cardiologent}.} \textit{Window level:} an agent per
signal reads the waveform through its tools---seven on the ECG, nine on the
PPG---and returns a grounded class, and
window fusion reconciles the ECG and PPG readings by debate into one verdict.
\textit{Patient level:} the verdicts are aggregated into a rhythm profile,
reconciled across signals by the same debate, and only then read against the
patient's metadata through guideline retrieval, a reasoning stage, and a critic,
yielding the integrated diagnosis, its significance, and the urgency and
management.}
\label{fig:model}
\end{figure*}
 
\section{Dataset}
 
The task we formulate calls for data of a particular kind: two signals recorded
at once, continuously, over a monitoring procedure, with the rhythm in them
annotated. VitalDB is an open database of intraoperative monitoring from 6,388 non-cardiac
surgical cases, recording ECG and PPG concurrently alongside each patient's
clinical record but saying nothing about the rhythm they
contain~\citep{lee2022vitaldb}; the VitalDB Arrhythmia Database supplies that,
annotating 734,528 seconds of that ECG from 482 of the patients with 661,894
heartbeats across four beat types and ten rhythm categories, read by at least
two of five anesthesiologists at $\kappa = 0.930$~\citep{eun2026vitaldbarr}.
 
Of the 482 annotated patients, 426 have both signals, giving 166,026 windows.
Because the labels are on the ECG, we transplant them onto the PPG, aligning
each R~peak with the pulse it drove by pulse arrival time. On the ECG a window
carries one of nine classes---\emph{normal}, atrial fibrillation, multifocal
atrial tachycardia, supraventricular tachycardia, ventricular tachycardia, PVC,
PAC, AV block, and sinus node dysfunction. On the PPG these collapse to a coarser
seven, since the pulse cannot resolve some distinctions the ECG can; the merged
set and its rationale are given with the PPG agent (Method).
 
The split is by patient, so no case contributes windows to more than one side of
it: 302 patients and 126,959 windows train the specialists, 59 patients and
20,194 windows are held out for validation, and the remaining 65 patients and
18,873 windows are the test set, on which every result reported here is
computed.

\section{Problem Formulation}
 
A monitored record consists of two synchronized signals---a lead~II ECG and a
PPG---acquired continuously through a procedure, together with the patient's own
data: age, sex, comorbidity, and the rest of the case record. The record is divided into short windows, each spanning a fixed number of beats (around ten seconds; Dataset).
 
\textbf{Window level.} Each window is assigned one arrhythmia class---nine when both 
the ECG and the PPG are present, and a coarser seven in the PPG-only setting
(Dataset)---together with the span of signal that is abnormal,
which the class alone does not give.
 
\textbf{Patient level.} Given every window of a record and the patient's own
data, the task is to say what the record means for that patient, in three parts.
Which arrhythmias appear, where, and which is the dominant rhythm---a property
of the record and of no window in it. What that profile signifies for this
patient, and which single finding is the one to act on. And what should be done:
further monitoring or intervention, at what urgency, on what evidence.

\section{Method}
 
\textsc{Cardiologent} works at the two levels defined above, shown in
Figure~\ref{fig:model}. At the window
level, an agent per signal reads the ECG or the PPG and the two readings are
reconciled into one verdict per window. At the patient level, each agent's window readings are assembled across the record into a rhythm profile, the two signals'
profiles are reconciled, and only afterwards is the profile read against the
patient's own data and the clinical guidelines. Both levels reconcile the two
signals by the same debate, described under window fusion below; we take the
pipeline in order.
 
\subsection{ECG and PPG Agents}
 
Each signal is read by its own agent, one for the ECG and one for the PPG,
identical in construction. At the centre of each is a \emph{specialist}: a LLaVA-7B multimodal
model~\citep{liu2023llava} fine-tuned to read the raw waveform and its plot.
A language model reads a raw physiological trace only weakly on its
own~\citep{zhou2023onefitsall,jin2024timellmtimeseriesforecasting,alnegheimish2024sigllm,zhou2025tsanomalies,quinlan2026chatts},
so we adapt it to the signal, as recent ECG language models
do~\citep{lan2025gem,liu2026pulse,jin2026ecgr1}. Around it sit
deterministic tools that measure the quantities a rhythm reading turns
on---rate, R--R regularity, and P-wave and QRS descriptors on the ECG, pulse-wave
morphology and perfusion on the PPG---so every number the agent reports is a
measured value rather than one read off the trace. Each agent reaches its reading through its own internal deliberation (below) and returns one reading: a class, the abnormal span, and a rationale citing the measurements behind it. Because each agent is complete on its own, the system
matches whatever a deployment provides: the PPG agent alone in a wearable, both
together where the ECG is monitored too. Running alone costs resolution---the
pulse cannot separate ventricular from atrial ectopy, nor MAT from
VT~\citep{page2016svt,alkhatib2018va,han2020pacpvc}, so
PVC merges with PAC and VT with MAT, leaving the coarser seven. Architecture and
prompts are in the supplement.
 
\subsection{Window Fusion}
 
The two readings are reconciled into one verdict by a \emph{debate}, several
instances arguing the reading out across rounds rather than one model refining
its own answer, which tends to defend a confident
error~\citep{du2024debate,liang2024mad}. Because the agents use different label
sets---nine classes on the ECG, seven on the PPG with ectopy and
VT/MAT merged where the pulse cannot separate them---the
readings are first checked for compatibility, telling one finding named two ways
from a real conflict, then argued through three roles:
\begin{itemize}
\item \textbf{Analysts} re-examine both readings, each along one axis---rate,
atrial activity, and ventricular morphology---weighing each modality by how well
its call is grounded in its own measured evidence rather than by a fixed
precedence, since neither signal is uniformly the better witness.
 
\item A \textbf{devil's advocate} then argues against the emerging verdict,
pressing its weakest points; this adversarial role is what separates debate from
self-refinement, and keeps a confident but wrong reading from passing
unchallenged~\citep{kim2024debate}. Whereas that work uses the devil's advocate to
\emph{score} generated text, here it contests a diagnostic decision and forces the
analysts to re-examine the measured evidence behind it. The analysts answer it in
a further round.
 
\item A \textbf{referee} weighs the exchange and issues the verdict, so a correct
minority reading can prevail on its evidence rather than be
outvoted~\citep{du2024debate}. It grants authority feature by feature to the
signal that can see the feature: the ECG's QRS width and P~wave---which the pulse
cannot resolve---split a merged PPG class such as VT/MAT into
VT or a supraventricular rhythm, while the PPG's pulse regularity and
perfusion carry the haemodynamic significance the ECG does not show.
\end{itemize}
 
The verdict is one nine-class label; the abnormal span is the union of the two.
In the wearable scenario the PPG agent's own verdict stands.
 
\subsection{Patient-Level Assembly}
 
\begin{itemize}
\item \textbf{Aggregation.} Each agent's window readings are gathered, in time
order, into a record-level profile per signal---which arrhythmias occur, over how much of
the record, and which dominates---that no single window contains; the PPG profile
is held to what the pulse can support, flagging rather than asserting an atrial
rhythm.
 
\item \textbf{Patient fusion.} The ECG and PPG profiles are reconciled by the
same window-fusion debate, judged now on authority and severity rather than
per-feature evidence: authority weighs how confident each profile is, so the ECG
leads where it is sure and yields where it is not, and severity breaks an even
exchange toward the more dangerous reading, so the finding a clinician must not
miss survives.
\end{itemize}
 
\subsection{Patient Reasoning}
 
Only once the profile is fixed does the patient enter, and the remaining stages
turn the finding into a decision.
 
\begin{itemize}
\item \textbf{Retrieval (RAG).} The profile and the metadata generate a few
clinical queries about the significance, management, and risk of the
finding---not its identity, which is already settled---and retrieve matching
passages from two sources: a curated corpus of arrhythmia guideline statements
and the PubMed literature. Each retrieved passage carries its citation, so the
evidence it supplies arrives already referenced; the corpus and retrieval details
are in the supplement.
 
\item \textbf{Reasoning.} A debate over the metadata, the profile, and the
retrieved passages produces the three patient-level outputs: the integrated
diagnosis, its clinical significance, and the urgency with the management, each
tied to a retrieved passage and citing the PubMed reference that passage carries.
 
\item \textbf{Critic.} A final pass checks the assessment against those
passages---that each claim is grounded in the profile and supported by a citation
actually retrieved, not one invented---and revises it where it is not, before the
outputs are returned.
\end{itemize}
 
The order is deliberate: the record is read and reconciled before the patient is
seen, so that what the arrhythmia \emph{is} rests on the waveform, and only what
it \emph{means} draws on who the patient is.

 
\section{Experiments}
 
We evaluate \textsc{Cardiologent} at two levels---the window level and the
patient level---against one fixed set of
baselines: five general-purpose vision--language models
(GPT-4o~\citep{hurst2024gpt4o}, Gemini~2.5~Pro~\citep{comanici2025gemini25},
Qwen2.5-VL-72B~\citep{bai2025qwen25vl},
Llama-3.2-90B-Vision,
MedGemma-27B~\citep{sellergren2025medgemma}) given identical inputs and questions.
The baselines are prompted directly at both levels, so the gap at each measures
what the construction adds over a capable model given the same inputs. Evaluation is on $3{,}000$ windows
sampled from the held-out test patients, covering every test patient and
preserving class proportions. Reasoning is scored by an LLM
judge~\citep{zheng2023judge,gu2024judgesurvey}---DeepSeek-V4-Pro, outside the
baseline set, so no system grades its own outputs, and one
whose scores agree with blinded cardiologists as closely as
the cardiologists agree with each other (patient level; Figure~\ref{fig:icc}). The orchestration (analysts, devil’s advocate, referee,
and patient-level stages) runs on Qwen2.5-72B, an openweight
model, so the whole pipeline can be run on-premises
without sending patient data to an external API; since one
baseline is the vision variant of that same model at the same
scale, the gap over it isolates the agent construction from the
backbone. Rubrics and sampling are in the supplement.

\paragraph{Metrics.} Classification is reported as macro-F1 and balanced
accuracy, not inflated by the dominant \emph{normal} class. Grounding uses IoU
and, since a whole-window span scores a high IoU without localizing, a per-sample
MCC~\citep{matthews1975mcc,chicco2020mcc} we treat as primary. Reasoning is judged
on four axes, each $1$--$4$, on every window probe:
\begin{itemize}
\item \textbf{Correctness} --- is the diagnosis itself right (the rhythm matches
ground truth)?
\item \textbf{Faithfulness} --- are the features it cites real, with measured
values that match the signal rather than fabricated?
\item \textbf{Validity} --- do those features actually support the diagnosis, or
are they irrelevant to it?
\item \textbf{Completeness} --- are the discriminators that separate this rhythm
from its neighbors all addressed, or is the reasoning partial?
\end{itemize}
The window agent is drawn out through three probes:
\begin{itemize}
\item \textbf{T1} --- diagnosis with supporting features.
\item \textbf{T2} --- differential, excluding each alternative.
\item \textbf{T3} --- the one change that would alter the diagnosis.
\end{itemize}
Each probe is scored by the judge against a per-window gold reference anchored to
the ground-truth label and cross-checked by a separate model (supplement).
The patient pipeline is scored on three axes, each two sub-axes ($0$--$4$, to $8$):
\begin{itemize}
\item \textbf{P1} --- integrated diagnosis: \emph{correctness} (the dominant
rhythm matches ground truth and secondary arrhythmias are named) and
\emph{faithfulness} (every claim traces back to the waveform).
\item \textbf{P2} --- clinical significance: \emph{prioritization} (the single
most dangerous finding is foregrounded) and \emph{context} (age, comorbidity, and
surgical setting are used to drive the risk).
\item \textbf{P3} --- urgency and management: \emph{management} (a specific,
guideline-concordant recommendation tied to the class and patient) and
\emph{urgency and safety} (the stated urgency is right and the plan is safe).
\end{itemize}
Per-grade rubrics are in the supplement.
 
\subsection{\textsc{Cardiologent} at the Window Level}
 
The window level wraps the two specialists in the full system---deterministic
tools, the internal debate, and window fusion across the two signals---and reads
out a per-window verdict. The specialists are also measured on their own signal in
the supplement, where each, at $7$B, leads every baseline on both classification
and reasoning.
 
\paragraph{Classification.} \textsc{Cardiologent} raises nine/seven-class macro-F1
to $0.449$ (both) and $0.425$ (PPG), roughly double the strongest baseline
(Table~\ref{tab:agent-cls}). As a binary normal-vs-arrhythmia decision it reaches
balanced accuracy above $0.9$ in both settings, while most baselines collapse
toward chance from one side, calling nearly all windows normal or nearly all
abnormal. GPT-4o is the apparent exception at $0.83$ on \emph{both}, but its low
macro-F1 shows this rests on the two frequent classes: it separates AF from normal,
misses the rarer rhythms, and falls to $0.56$ once the ECG is removed. Detection is
the easier tier; the nine-class macro-F1 above it is where the gap is widest.
 
\paragraph{Reasoning.} The window agent is probed on diagnosis (T1), differential
(T2), and the one change that would alter the call (T3); \textsc{Cardiologent}
leads all six task--setting combinations with non-overlapping confidence intervals
(Table~\ref{tab:agent-reason}), by a similar margin whether both signals or the
PPG alone is read. Absolute scores run higher under PPG, but this reflects its
coarser label space and the correspondingly relaxed judging, so we read the gap to
the best baseline, stable across settings, not the raw level. T2 is our hardest
probe---ruling alternatives out is harder than naming a rhythm---and the smaller
baselines fall below $6$ there, listing possibilities without excluding them,
while we tie each exclusion to a measured feature. T3, counterfactual reasoning, is
where the general models are most competitive and our margin is narrowest.
 
\paragraph{Grounding.} Grounding is scored only on focal rhythms (PVC, PAC, VT,
SVT), where the abnormality occupies part of the window; for sustained rhythms any
span scores trivially. On overlap alone (focal-IoU) the baselines look
competitive, but overlap rewards marking the whole window, so we take the
per-sample focal-MCC as primary. There the separation is stark
(Table~\ref{tab:ground}): \textsc{Cardiologent} reaches $0.305$ (both) and $0.319$
(PPG), while every general model sits at or below $0.18$---most within noise of
zero, MedGemma negative---despite respectable IoU, shading the window rather than
localizing the abnormality.
 
\begin{table}[t]
\centering\small
\caption{\textsc{Cardiologent} \textbf{window-level} classification: nine/seven-class
macro-F1 (top) and binary normal-vs-arrhythmia balanced accuracy (bottom). $\pm$ is
a $95\%$ bootstrap CI.}
\label{tab:agent-cls}
\setlength{\tabcolsep}{5pt}
\begin{tabular}{lcc}
\toprule
& both & PPG \\
\midrule
\multicolumn{3}{l}{\emph{macro-F1}}\\
\textbf{\textsc{Cardiologent} (Ours)} & $\mathbf{0.449}{\scriptstyle\pm0.026}$ & $\mathbf{0.425}{\scriptstyle\pm0.027}$ \\
GPT-4o         & $0.255{\scriptstyle\pm0.045}$ & $0.147{\scriptstyle\pm0.024}$ \\
Gemini-2.5-Pro & $0.186{\scriptstyle\pm0.042}$ & $0.194{\scriptstyle\pm0.020}$ \\
Qwen2.5-VL-72B & $0.160{\scriptstyle\pm0.015}$ & $0.198{\scriptstyle\pm0.011}$ \\
MedGemma-27B   & $0.084{\scriptstyle\pm0.002}$ & $0.131{\scriptstyle\pm0.011}$ \\
Llama-3.2-90B-Vision  & $0.021{\scriptstyle\pm0.004}$ & $0.169{\scriptstyle\pm0.003}$ \\
\midrule
\multicolumn{3}{l}{\emph{binary accuracy}}\\
\textbf{\textsc{Cardiologent} (Ours)} & $\mathbf{0.94}{\scriptstyle\pm0.01}$ & $\mathbf{0.92}{\scriptstyle\pm0.01}$ \\
GPT-4o         & $0.83{\scriptstyle\pm0.01}$ & $0.56{\scriptstyle\pm0.01}$ \\
Gemini-2.5-Pro & $0.61{\scriptstyle\pm0.01}$ & $0.61{\scriptstyle\pm0.01}$ \\
Qwen2.5-VL-72B & $0.58{\scriptstyle\pm0.01}$ & $0.53{\scriptstyle\pm0.01}$ \\
MedGemma-27B   & $0.54{\scriptstyle\pm0.01}$ & $0.53{\scriptstyle\pm0.01}$ \\
Llama-3.2-90B-Vision  & $0.50{\scriptstyle\pm0.00}$ & $0.50{\scriptstyle\pm0.00}$ \\
\bottomrule
\end{tabular}
\end{table}
 
\begin{table}[t]
\centering\small
\caption{\textsc{Cardiologent} \textbf{window-level} reasoning per probe (SUM$/16$).
$\pm$ is a $95\%$ bootstrap CI.}
\label{tab:agent-reason}
\setlength{\tabcolsep}{4pt}
\begin{tabular}{lccc}
\toprule
& T1 & T2 & T3 \\
\midrule
\multicolumn{4}{l}{\emph{both (ECG+PPG)}}\\
\textbf{\textsc{Cardiologent} (Ours)} & $\mathbf{12.51}{\scriptstyle\pm0.11}$ & $\mathbf{10.50}{\scriptstyle\pm0.17}$ & $\mathbf{12.88}{\scriptstyle\pm0.19}$ \\
GPT-4o         & $11.57{\scriptstyle\pm0.16}$ & $8.99{\scriptstyle\pm0.20}$ & $12.02{\scriptstyle\pm0.19}$ \\
Gemini-2.5-Pro & $8.66{\scriptstyle\pm0.16}$ & $9.94{\scriptstyle\pm0.24}$ & $11.30{\scriptstyle\pm0.21}$ \\
Qwen2.5-VL-72B & $10.74{\scriptstyle\pm0.18}$ & $5.55{\scriptstyle\pm0.14}$ & $8.21{\scriptstyle\pm0.22}$ \\
MedGemma-27B   & $9.75{\scriptstyle\pm0.16}$ & $4.87{\scriptstyle\pm0.09}$ & $7.21{\scriptstyle\pm0.19}$ \\
Llama-3.2-90B-Vision  & $4.47{\scriptstyle\pm0.06}$ & $4.78{\scriptstyle\pm0.09}$ & $7.58{\scriptstyle\pm0.20}$ \\
\midrule
\multicolumn{4}{l}{\emph{PPG-only}}\\
\textbf{\textsc{Cardiologent} (Ours)} & $\mathbf{12.90}{\scriptstyle\pm0.12}$ & $\mathbf{8.26}{\scriptstyle\pm0.13}$ & $\mathbf{12.19}{\scriptstyle\pm0.21}$ \\
GPT-4o         & $8.21{\scriptstyle\pm0.18}$ & $7.01{\scriptstyle\pm0.09}$ & $11.30{\scriptstyle\pm0.16}$ \\
Gemini-2.5-Pro & $10.50{\scriptstyle\pm0.17}$ & $6.56{\scriptstyle\pm0.16}$ & $9.29{\scriptstyle\pm0.17}$ \\
Qwen2.5-VL-72B & $11.03{\scriptstyle\pm0.18}$ & $6.65{\scriptstyle\pm0.11}$ & $7.63{\scriptstyle\pm0.14}$ \\
MedGemma-27B   & $10.60{\scriptstyle\pm0.18}$ & $5.78{\scriptstyle\pm0.09}$ & $7.22{\scriptstyle\pm0.13}$ \\
Llama-3.2-90B-Vision  & $5.80{\scriptstyle\pm0.11}$ & $7.07{\scriptstyle\pm0.11}$ & $8.38{\scriptstyle\pm0.18}$ \\
\bottomrule
\end{tabular}
\end{table}
 
\begin{table}[t]
\centering\small
\caption{\textsc{Cardiologent} \textbf{window-level grounding} on focal rhythms. MCC is primary; IoU
alone rewards marking the whole window. both $n{=}230$, PPG $n{=}209$. $\pm$ is a
$95\%$ bootstrap CI.}
\label{tab:ground}
\setlength{\tabcolsep}{3.5pt}
\begin{tabular}{lcccc}
\toprule
& \multicolumn{2}{c}{both} & \multicolumn{2}{c}{PPG} \\
\cmidrule(lr){2-3}\cmidrule(lr){4-5}
& IoU & MCC & IoU & MCC \\
\midrule
\textbf{\textsc{Cardiologent} (Ours)} & \textbf{0.584} & \textbf{0.305} & \textbf{0.569} & \textbf{0.319} \\
Llama-3.2-90B-Vision  & 0.533 & 0.001 & 0.544 & 0.000 \\
GPT-4o         & 0.489 & 0.005 & 0.357 & 0.029 \\
Gemini-2.5-Pro & 0.440 & 0.178 & 0.535 & 0.109 \\
Qwen2.5-VL-72B & 0.292 & 0.056 & 0.433 & 0.037 \\
MedGemma-27B   & 0.230 & -0.006 & 0.182 & -0.011 \\
\bottomrule
\end{tabular}
\end{table}
 
\subsection{\textsc{Cardiologent} at the Patient Level}
 
\begin{figure}[t]
\centering
\includegraphics[width=0.82\linewidth]{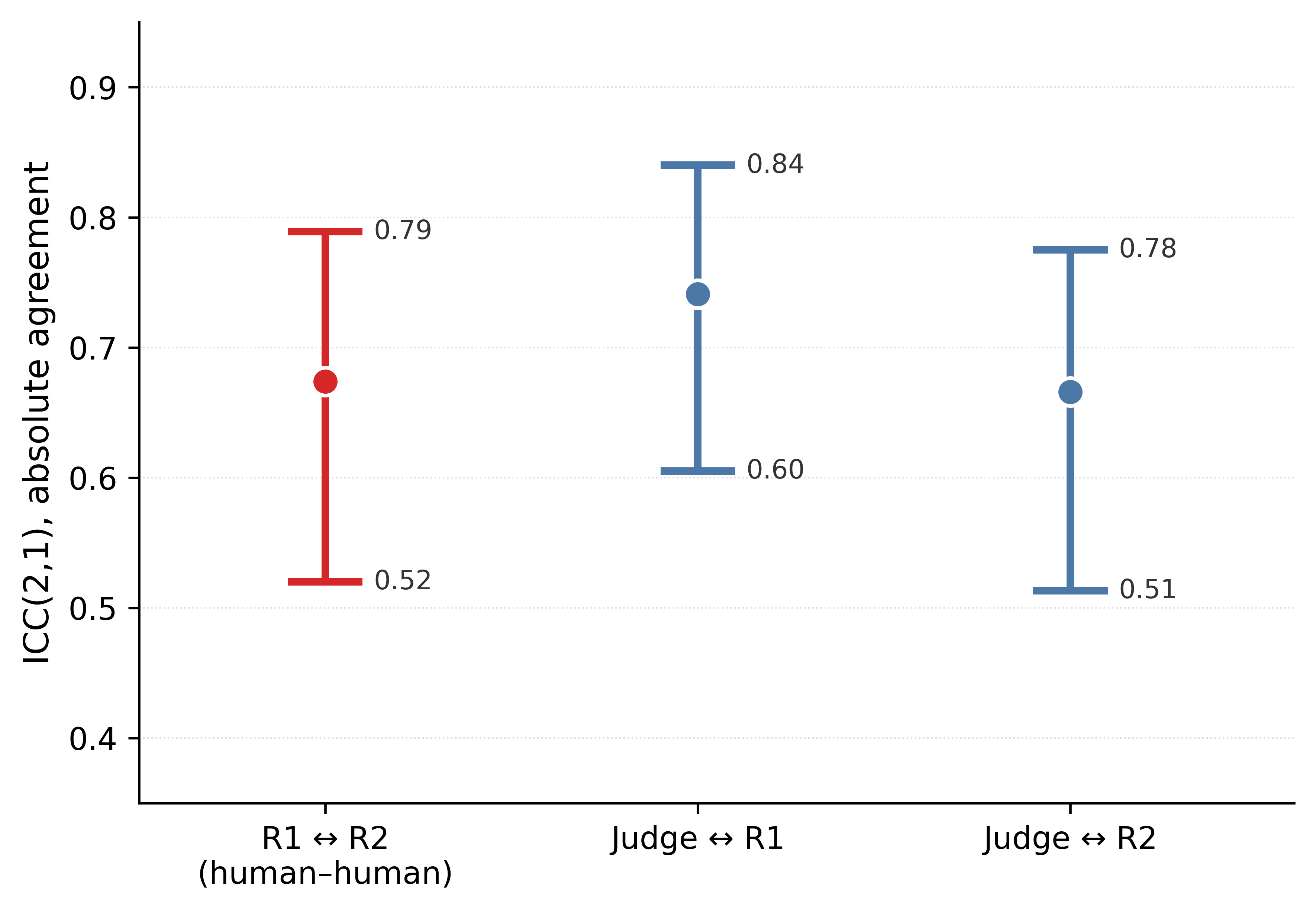}
\caption{\textbf{Rater agreement} (ICC(2,1), absolute agreement, with $95\%$ CI)
on the $20$-patient subset. The judge's agreement with each cardiologist
($0.74$, $0.66$) is comparable to the two cardiologists' agreement with each other
($0.67$): the judge agrees with a cardiologist about as closely as the two
cardiologists agree with each other.}
\label{fig:icc}
\end{figure}

The patient level assembles the window verdicts, with the patient's metadata and
retrieved guidelines, into the three-part assessment; the baselines are asked the
same three questions on the same record. \textsc{Cardiologent} leads on all three
(Table~\ref{tab:patient}), by the widest margin on P1, the integrated diagnosis
the assembly is built to reach. Two cardiologists blind to system identity rank
the systems in the judge's order and place \textsc{Cardiologent} first throughout;
their agreement with the judge (ICC $0.74$, $0.66$) matches their agreement with
each other ($0.67$, Figure~\ref{fig:icc}), so the at-scale judge stands in for a
clinician about as well as one clinician stands in for another.
 
We also score the diagnosis directly---the dominant rhythm named exactly, within
its management family, or anywhere at all (supplement)---where
\textsc{Cardiologent} leads at every grade and the baselines exceed a third only
at the loosest.
 
\begin{table}[t]
\centering\small
\caption{\textbf{Patient-level} assessment on the three tasks, each scored on two
axes (SUM$/8$). Top: the LLM judge (DeepSeek-V4-Pro) over all $65$ test patients;
bottom: two cardiologists, blind to system identity, over a $20$-patient subset
drawn at random but stratified to preserve the arrhythmia-class proportions of
the full test set (per-class counts in the supplement).
$\pm$ is a $95\%$ cluster bootstrap CI ($B{=}10{,}000$).}
\label{tab:patient}
\setlength{\tabcolsep}{6pt}
\begin{tabular}{lccc}
\toprule
& P1 & P2 & P3 \\
\midrule
\multicolumn{4}{@{}l}{\emph{LLM judge} ($n{=}65$)}\\
\textbf{\textsc{Cardiologent} (Ours)} & $\mathbf{4.66}{\scriptstyle\pm0.61}$ & $\mathbf{4.37}{\scriptstyle\pm0.64}$ & $\mathbf{4.55}{\scriptstyle\pm0.58}$ \\
Gemini-2.5-Pro & $3.20{\scriptstyle\pm0.50}$ & $3.31{\scriptstyle\pm0.59}$ & $3.94{\scriptstyle\pm0.54}$ \\
GPT-4o         & $2.78{\scriptstyle\pm0.38}$ & $2.83{\scriptstyle\pm0.42}$ & $2.94{\scriptstyle\pm0.36}$ \\
Qwen2.5-VL-72B & $2.12{\scriptstyle\pm0.26}$ & $2.23{\scriptstyle\pm0.21}$ & $2.86{\scriptstyle\pm0.22}$ \\
MedGemma-27B   & $1.91{\scriptstyle\pm0.07}$ & $2.09{\scriptstyle\pm0.23}$ & $2.89{\scriptstyle\pm0.21}$ \\
Llama-3.2-90B-Vision  & $2.06{\scriptstyle\pm0.20}$ & $1.49{\scriptstyle\pm0.29}$ & $2.91{\scriptstyle\pm0.32}$ \\
\midrule
\multicolumn{4}{@{}l}{\emph{Cardiologists} ($n{=}20$, two raters)}\\
\textbf{\textsc{Cardiologent} (Ours)} & $\mathbf{4.58}{\scriptstyle\pm1.32}$ & $\mathbf{4.70}{\scriptstyle\pm1.08}$ & $\mathbf{5.03}{\scriptstyle\pm1.01}$ \\
Gemini-2.5-Pro & $3.33{\scriptstyle\pm1.36}$ & $3.75{\scriptstyle\pm1.09}$ & $4.22{\scriptstyle\pm0.99}$ \\
GPT-4o         & $2.98{\scriptstyle\pm1.11}$ & $3.48{\scriptstyle\pm0.88}$ & $3.85{\scriptstyle\pm0.95}$ \\
Qwen2.5-VL-72B & $1.48{\scriptstyle\pm0.78}$ & $2.48{\scriptstyle\pm0.54}$ & $2.92{\scriptstyle\pm0.45}$ \\
MedGemma-27B   & $1.38{\scriptstyle\pm0.60}$ & $2.62{\scriptstyle\pm0.54}$ & $2.80{\scriptstyle\pm0.47}$ \\
Llama-3.2-90B-Vision  & $1.00{\scriptstyle\pm0.35}$ & $1.88{\scriptstyle\pm0.21}$ & $2.25{\scriptstyle\pm0.31}$ \\
\bottomrule
\end{tabular}
\end{table}
 
\section{Limitations}
 
This work has several limitations. Our results come from one institution's
recordings, so the class balance, the patient mix, and the recording conditions
are particular to it. Coverage is thinnest on the rare classes---SVT, VT, AV
block---which hold the macro-F1 down and are the rhythms it would matter most to
catch. The PPG-only setting has never been tested on a wearable: our pulse comes
from a monitor, on a still patient with a fixed sensor. The evaluation is also
retrospective---we score the decision itself, not whether acting on it would have
changed the patient's course---and the two deployment states are evaluated
separately rather than as one pathway: every record carries both signals, so the
hand-off from pulse-only surveillance to a combined read is never exercised.
 
\section{Conclusion and Future Work}
 
We formulate patient-level arrhythmia decision support as a task---what
arrhythmia a monitored record holds, what it means for this patient, and what
should be done about it---and present \textsc{Cardiologent}, a multi-agent system
that spans it from the window reading to the decision, with each conclusion held
to the guideline it cites. Across integrated diagnosis, clinical significance,
and urgency and management, \textsc{Cardiologent} scores highest on every
axis---under cardiologists and under the LLM judge alike---and its window-level
macro-F1 ($0.449$) is roughly double the strongest baseline's. We hope a decision a clinician can audit, rather than one more unexplained alert, is a step toward systems usable in continuous monitoring.
 
The limitations above each point to a next step: external validation on multi-centre data;
targeted collection of the high-risk rhythms rather than
more of the same recordings; validation on ambulatory recordings; the pathway
itself, evaluated end to end---a wearable running the PPG agent alone for
continuous, low-cost surveillance, and, on a finding that warrants it, the
patient moving to a monitored setting where the ECG is added and the two agents
combine, raising the level of scrutiny without changing systems; a retrospective
study on a hospital's own monitoring data; and, above all,
placing the system in front of clinicians prospectively, to measure whether it
changes what they do and to what end.
 
\bibliography{reference}
 
\newtcolorbox{rubricbox}[1]{
  enhanced, breakable, sharp corners,
  colback=black!3, colframe=black!80,
  boxrule=0.4pt, left=6pt, right=6pt, top=4pt, bottom=4pt,
  title={#1}, fonttitle=\bfseries\small, coltitle=white,
  colbacktitle=black!85,
  toptitle=2pt, bottomtitle=2pt, lefttitle=6pt, righttitle=6pt
}
\newcommand{\rubtask}[1]{\smallskip\noindent\textbf{\itshape #1}\par\nopagebreak\vspace{2pt}}
\newcommand{\gradeline}[2]{%
  \par\noindent\hangindent=1.5em\hangafter=0
  \makebox[1.5em][l]{\textbf{#1}}#2}
\newcommand{\rubaxis}[5]{%
  \par\smallskip\noindent\textbf{#1}%
  \gradeline{4}{#2}\gradeline{3}{#3}\gradeline{2}{#4}\gradeline{1}{#5}\par}
\newcommand{\rubaxisF}[6]{%
  \par\smallskip\noindent\textbf{#1}%
  \gradeline{4}{#2}\gradeline{3}{#3}\gradeline{2}{#4}\gradeline{1}{#5}\gradeline{0}{#6}\par}
\newtcolorbox{promptbox}[1]{
  enhanced, breakable, sharp corners,
  colback=black!3, colframe=black!80,
  boxrule=0.4pt, left=6pt, right=6pt, top=4pt, bottom=4pt,
  title={#1}, fonttitle=\bfseries\small, coltitle=white,
  colbacktitle=black!85,
  toptitle=2pt, bottomtitle=2pt, lefttitle=6pt, righttitle=6pt
}
\newcommand{\prole}[1]{\smallskip\noindent\textbf{#1}\ }
 
%
 
\clearpage 
\appendix

\section{Related Work}
 
\paragraph{Language models for time series.}
Whether a language model can read a raw numerical signal has been asked in a
fairly clear progression. Early work reprogrammed a frozen model for forecasting
and classification, showing a general backbone could be adapted to time series at
all~\citep{zhou2023onefitsall,jin2024timellmtimeseriesforecasting}. Attention
then moved to anomaly detection, where a model was prompted zero-shot to flag the
abnormal part of a series~\citep{alnegheimish2024sigllm}; the limits of that
prompting were examined soon after, finding the ability shallow and
inconsistent~\citep{zhou2025tsanomalies}, and more recent systems pair the signal
with text for multimodal reasoning rather than raw
prompting~\citep{quinlan2026chatts}. The throughline is that a general model reads
a raw trace only weakly, which is why our specialists are adapted to the waveform
rather than prompted on it; and unlike this line, the signal here is not the end
product but a tool the agent calls.
 
\paragraph{ECG and signal language models.}
Attaching a language model to the ECG began with instruction tuning for report
generation and question answering, which gave a waveform a textual
reading~\citep{wan2025meit,zhao2025ecgchat,yang2025ecglm}. Attention then turned
to making that reading trustworthy rather than fluent: grounding each statement
in the time series it refers to~\citep{lan2025gem}, following an explicit
diagnostic protocol~\citep{jin2026ecgr1}, and unifying understanding with
generation in one model~\citep{jin2025uniecg}; a parallel thread pretrained ECG
foundation models at
scale~\citep{li2025ecgfounder,li2026anyecgchat}. Most
recently the reading has been wrapped in agents.
Zodiac~\citep{zhou2024zodiac} assembles a patient-specific report from a
monitoring study under cardiologist validation, but its arrhythmias arrive
already detected---a representative strip and a burden table selected
upstream---and it stops at interpretation, leaving the management decision to the
reviewing cardiologist; its validation scores the report---accuracy against the
patient's own data, completeness, organization, succinctness---not the diagnosis
against a label. CardAIc-Agents~\citep{zhang2025cardaic} instead ingests the raw
signal and coordinates tools over it, but its task is disease-level diagnosis
(heart failure, myocardial infarction) scored by classification accuracy, not the
rhythm-level identification of arrhythmias, and its own signal-level wave
detection is reported as suboptimal; and while its intermediate visual outputs are
reviewed by cardiologists, the management it plans is not among what is scored. A tool-calling dialogue agent~\citep{chung2026ecgagent} reads a single
ECG over a short exchange. Across these, the object is one recording or one
disease label, returned as a reading or a report---none assembles a monitored
record, window by window, into a patient-level arrhythmia profile and carries it
through to an audited management decision. Our specialist sits at the end of
this line, but the system around it does what the line has not: it reconciles a
second and weaker signal against the ECG, and it carries the per-window reading
up into a patient-level assessment.
 
\paragraph{LLM agents.}
Beyond a single model, agentic systems let a language model plan, call tools,
and revise its answer, and multi-agent debate improves reasoning by having
several instances argue a question to an adjudicated
verdict~\citep{du2024debate,liang2024mad,kim2024debate}. Our fusion is an
instance of this pattern, but with a difference the setting forces: the debaters
are not interchangeable and consensus is not the goal. Because the ECG and the
PPG see physically different things, authority is granted feature by feature to
the signal that can see a feature, and a referee adjudicates rather than a
majority---so a correct minority reading is not voted down.
 
\paragraph{Medical multi-agent systems.}
As agents matured, the pattern was carried into medicine, first by wrapping
individual clinical tools in a single tool-calling agent~\citep{li2024mmedagent},
then by coordinating several specialized tools under a reasoning loop for a given
modality~\citep{fallahpour2025medrax}, and more recently by adapting the
collaboration itself---varying structure with case
difficulty~\citep{kim2024mdagents} or specializing agents to a signal such as the
ECG~\citep{chung2026ecgagent}. These share our use of a language-model agent over
clinical tools, but each stays within a single modality and a single
interaction---one image or one ECG, answered in a turn or a short dialogue. What
is specific here is the substrate and the target: a continuously monitored pair
of signals rather than one image or a single ECG, and an assessment assembled
window by window across a whole record into a diagnosis, its significance, and a
management decision.
 
\begin{table}[t]
\centering\small
\caption{Nine-class window counts over the full pool and the patient-disjoint
train/val/test split.}
\label{tab:classdist}
\begin{tabular}{lrrrr}
\toprule
class & total & train & val & test \\
\midrule
normal              & 75{,}039 & 51{,}827 & 12{,}069 & 11{,}143 \\
atrial fibrillation & 44{,}583 & 36{,}291 &  3{,}868 &  4{,}424 \\
PVC                 & 10{,}940 &  9{,}586 &    789 &    565 \\
PAC                 &  9{,}997 &  8{,}491 &    990 &    516 \\
MAT                 &  9{,}683 &  7{,}801 &    900 &    982 \\
SND                 &  6{,}493 &  5{,}312 &    648 &    533 \\
SVT                 &  5{,}215 &  4{,}383 &    515 &    317 \\
AVB                 &  3{,}528 &  2{,}819 &    361 &    348 \\
VT                  &    548 &    449 &     54 &     45 \\
\midrule
total               & 166{,}026 & 126{,}959 & 20{,}194 & 18{,}873 \\
\bottomrule
\end{tabular}
\end{table}
 
\begin{table}[t]
\centering\small
\caption{Class distribution of the $3{,}000$-window evaluation sample (ECG
nine-class).}
\label{tab:testdist}
\begin{tabular}{lrr}
\toprule
class & $n$ & \% \\
\midrule
normal              & 1{,}771 & 59.0 \\
atrial fibrillation &   703 & 23.4 \\
MAT                 &   156 &  5.2 \\
PVC                 &    90 &  3.0 \\
SND                 &    85 &  2.8 \\
PAC                 &    82 &  2.7 \\
AVB                 &    55 &  1.8 \\
SVT                 &    51 &  1.7 \\
VT                  &     7 &  0.2 \\
\midrule
total               & 3{,}000 & \\
\bottomrule
\end{tabular}
\end{table}
 
\section{Dataset Details}
 
A window spans twelve consecutive R-peaks rather than a fixed duration---one
leading, ten central, one trailing---which at $60$--$100$~bpm is roughly seven to
twelve seconds at $100$~Hz; windows are cut on the PPG peaks and the ECG is
aligned to them by the pulse transit time, the same offset that carries the
labels across (main text). The sliding stride is set per class to counteract
imbalance: the rare rhythms (SVT, VT, MAT, AVB) advance one peak at a time
($\sim$92\% overlap) so nearly every occurrence is kept, while common windows and
\emph{normal} advance three ($\sim$75\%) and are thereby subsampled. An episode is
never split across a boundary---a window either contains it in full or lies
wholly within it, and partial straddles are discarded---and a window is dropped
if any sample fails quality control (a PPG segment flagged by the signal-quality
index, an ECG beat marked bad, or a discard label in the mask) or if it holds
fewer than twelve annotated beats.
 
The resulting distribution is dominated by \emph{normal} and atrial fibrillation,
with the ventricular and conduction rhythms rare---the imbalance the
class-dependent stride is meant to soften. Table~\ref{tab:classdist} gives the
nine-class counts over the full pool and the patient-disjoint split, and
Table~\ref{tab:testdist} the $3{,}000$-window evaluation sample. The split is by
patient and stratified to preserve the overall class proportions across train,
validation, and test, so no patient appears on two sides and each split reflects
the same rhythm mix. The $3{,}000$-window evaluation sample is drawn from the
test set at random under two constraints---every test patient is represented and
the class proportions are preserved---so that it stays representative while
keeping the LLM-judge and API cost of scoring reasoning tractable.

\paragraph{Specialist QA.}
The specialist's Stage-2 targets---the two-to-three-sentence readings it is
trained to produce---are written once, offline, before any training. For each
window we render its plot with the detected R-peaks marked and pair it with two
text inputs: the window's arrhythmia annotation (the benchmark label, with
per-beat type, amplitude, QRS width, and R--R timing) as a ground-truth
\emph{anchor}, and a set of signal measurements summarizing rate and rhythm, the
rate profile, and per-interval morphology. A multimodal model (GPT-4o) reads the
plot and these inputs and writes an observation-to-evidence-to-diagnosis answer
whose conclusion matches the anchor and whose numbers are drawn from the supplied
measurements. The anchor guides accuracy but is instructed not to be copied
verbatim, so the target reads as a reading of the trace rather than a restatement
of the label.
 
\paragraph{Reasoning benchmark (T1/T2/T3).}
The reasoning probes are evaluated against a gold reference answer built per
window in a generate--review--regenerate pipeline. Each window carries up to three
fixed-template questions by its ground-truth class: T1 (diagnosis with supporting
features) on every window, T2 (differential, excluding each alternative) on every
arrhythmia, and T3 (the one change that would alter the call) on the
confusable classes (AF, MAT, PVC, PAC, VT, SVT); under PPG the P-wave and
QRS-morphology cues are replaced by pulse-wave morphology and the differential is
told to keep any ECG-only-separable pair merged. Golds are generated by GPT-4o
from four inputs---the ground-truth diagnosis as anchor, a short expert
description, the measured signal features, and the question---under the
constraint that the answer conclude with the anchor diagnosis and cite only
numbers present in those features, so the reference is grounded in the label rather
than in the generator's free recall. Every gold is then audited by a
\emph{different} model (Gemini-2.5-Pro) against the same anchor and evidence and
graded \textsc{correct}/\textsc{minor}/\textsc{incorrect}; the
\textsc{incorrect} cases are regenerated with the audit notes attached. Using one
model to write and a different one to check, with the ground-truth label---not
either model's opinion---as the standard, keeps the benchmark from being anchored
to a single model's judgement.
 
\section{Specialist Details}
 
Each specialist is trained in two stages (Figure~\ref{fig:specialist}). A
\emph{time-series encoder} is first pretrained on the raw waveform by masked
reconstruction, adapting the masked-autoencoder scheme~\citep{he2022mae} to the
signal: it is split into patches, a fraction are masked, and a Transformer
encoder--decoder reconstructs them, giving a representation of local morphology
without any labels. Once pretrained, the decoder is discarded and only the
encoder is kept, supplying the per-patch features used in Stage~2. The \emph{reasoning model}
is then a LLaVA-7B model~\citep{liu2023llava}---a Llama-7B backbone with a CLIP
vision encoder---that reads a plot of the window, encoded by CLIP, alongside a
text prompt carrying the measured features of the signal and the per-patch
features from the Stage-1 encoder. Presenting a raw signal together with text so
that a language model can reason over
both~\citep{alnegheimish2024sigllm,zhou2025tsanomalies,quinlan2026chatts} is the
setting this follows, here for the ECG and the pulse rather than a generic
series. It is fine-tuned to answer, in two or three sentences, whether the window
contains an arrhythmia and, if so, its class and the evidence.
 
\definecolor{phTaxonomy}{HTML}{D2691E}
\definecolor{phDetFeat}{HTML}{7B3FA0}
\definecolor{phPeak}{HTML}{E0A000}
\definecolor{phRaw}{HTML}{2E7D32}
\definecolor{phPatch}{HTML}{5B7B8A}
\definecolor{phReason}{HTML}{2AA5C0}
 
\begin{figure*}[t]
\centering
\includegraphics[width=\textwidth]{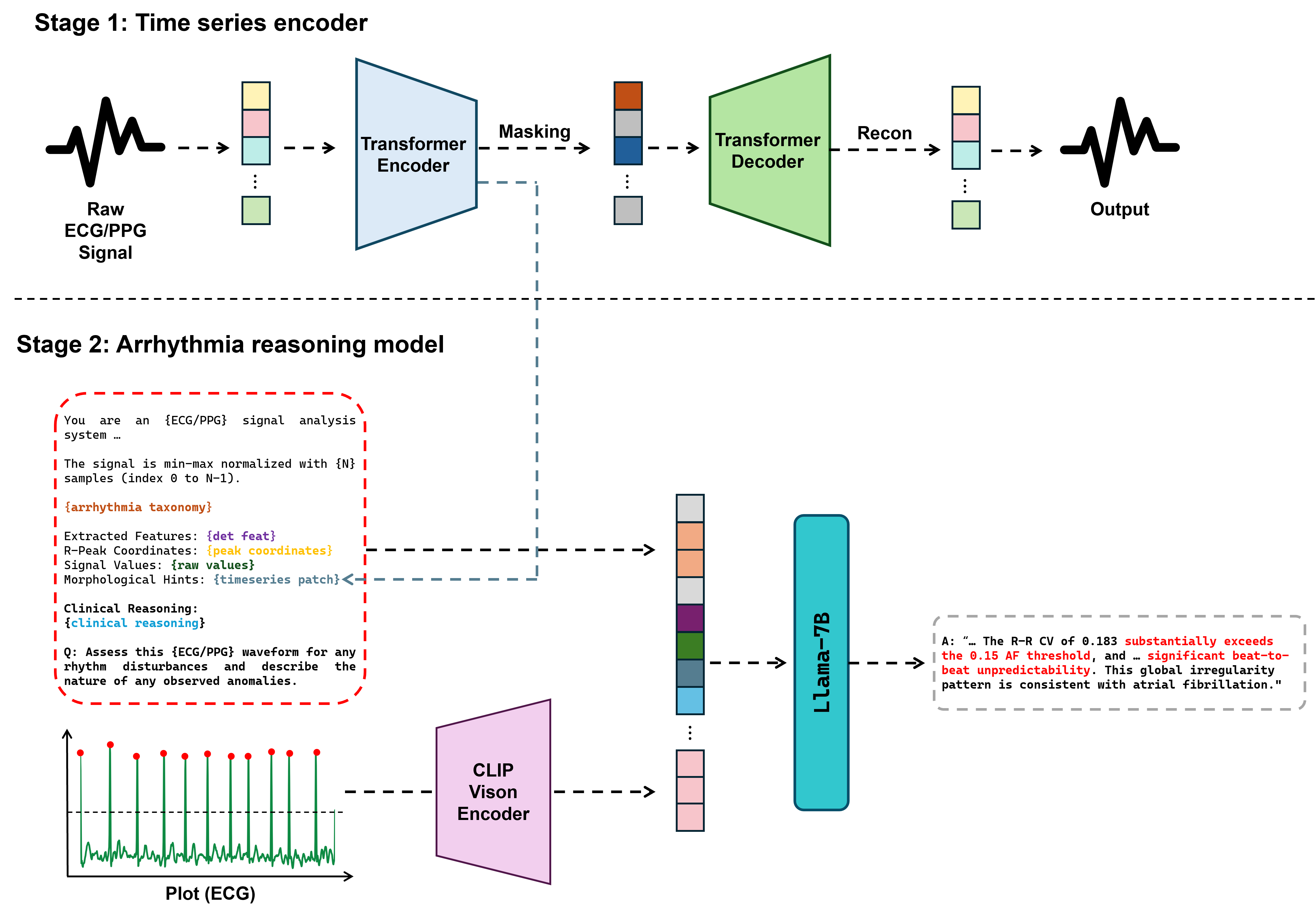}
\caption{The specialist. Stage~1 pretrains a time-series encoder by masked
reconstruction of the raw waveform. Stage~2 is the LLaVA-7B reasoning model
(Llama-7B with a CLIP vision encoder) reading a plot of the window together with
a prompt built from the measured features and the Stage-1 encoder, whose
per-patch features enter that prompt as
\textcolor{phPatch}{\textbf{\{timeseries patch\}}}.}
\label{fig:specialist}
\end{figure*}

The text prompt is where the grounding enters: every quantity in it is a measured
value or an encoder feature, not a number the model is left to read off the
trace. Its fields, named as in Figure~\ref{fig:specialist}, are:
\begin{itemize}
\item \textcolor{phTaxonomy}{\textbf{\{arrhythmia taxonomy\}}} --- the nine (ECG)
or seven (PPG) classes with their defining criteria, including the QRS-width and
P-wave distinctions the pulse cannot resolve.
\item \textcolor{phDetFeat}{\textbf{\{det feat\}}} --- the measured
features: beat count, heart rate, median R--R, R--R coefficient of variation,
maximum $|\Delta$R--R$|$, median QRS width, and premature-beat count.
\item \textcolor{phPeak}{\textbf{\{peak coordinates\}}} --- the detected R-peaks
as (index, amplitude).
\item \textcolor{phRaw}{\textbf{\{raw values\}}} --- the min--max normalized
waveform, subsampled.
\item \textcolor{phPatch}{\textbf{\{timeseries patch\}}} --- per-patch embeddings from the
Stage-1 encoder (the morphological hints).
\item \textcolor{phReason}{\textbf{\{clinical reasoning\}}} --- a scaffold laying
out the R--R, QRS-width, and amplitude sequences and walking four perspectives
(rhythm regularity, rate profile, QRS morphology, premature-beat morphology)
toward a conclusion.
\end{itemize}
The prompt closes with the question---assess the window and describe any anomaly.
On the PPG, which cannot resolve QRS width or premature-beat morphology, the QRS
and premature perspectives drop out and only rhythm regularity, rate, and
perfusion remain---the reason its taxonomy merges to seven classes.
 
\paragraph{Training objectives.}
Each stage optimizes a single loss term. Stage~1 is a peak-weighted masked
reconstruction: the loss is a mean-squared error computed only at masked
positions, upweighted around peaks. With $M_i$ the masked positions of sample
$i$ and $P_i$ the peaks detected on its target trace,
\begin{align*}
\mathcal{L}_1 &= \frac{1}{B}\sum_{i=1}^{B}
\operatorname{mean}_{t\in M_i}\!\Big[w_i(t)\,\big(\hat{x}_i(t)-x_i(t)\big)^2\Big],\\
w_i(t) &=
\begin{cases}5 & \text{if } \exists\,p\in P_i:\ |t-p|\le 10,\\ 1 & \text{otherwise.}\end{cases}
\end{align*}
Only masked positions contribute; predictions are interpolated back to each
sample's original resolution before the error is taken; and the peaks are found
on the target signal itself (local maxima above the mean, minimum spacing), not
from ground-truth annotations, so the objective needs no labels. Stage~2 is the
standard answer-only autoregressive cross-entropy,
\begin{equation*}
\mathcal{L}_2 = -\sum_{t\in A}\log p_\theta\!\left(y_t \mid y_{<t},\,\text{prompt},\,z_{\text{enc}}\right),
\end{equation*}
summed over the answer tokens $A$ (with the question and padding tokens masked
out). Neither stage uses an auxiliary term.
 
\paragraph{Training.}
Stage~1 pretrains the encoder as a TimeMAE on each window interpolated to $1024$
samples, with patch size $16$ ($64$ patches),
mask ratio $0.75$, and rotary position embeddings, over a $6$-layer
$8$-head Transformer ($d_{\text{model}}{=}512$); it is trained for $100$ epochs
(batch $4096$, learning rate $10^{-4}$). Stage~2 fine-tunes LLaVA-7B with a $64$-query
projector; only the Stage-1 signal encoder is frozen, and everything else---the
projector, the CLIP vision encoder, and the language backbone---is trained, for
$200$k steps (learning rate
$5{\times}10^{-6}$, bf16, FSDP with gradient checkpointing) on two A100-80GB GPUs.
 
\paragraph{Results.}
Each specialist is evaluated alone on its own signal---the ECG specialist on the
nine-class task, the PPG specialist on the seven-class merged task---against the
same five baselines, each given that signal, its plot, and the taxonomy. The written reading is
scored by the same judge on four axes ($1$--$4$: correctness, faithfulness,
validity, completeness; SUM$/16$).
At $7$B, several times smaller than every baseline, each specialist leads all of
them (Table~\ref{tab:spec-cls}): macro-F1 $0.393$ against $0.131$ on the ECG and
$0.367$ against $0.198$ on the PPG. Raw accuracy separates them far less
($0.762$ against Qwen's $0.587$), which is why we do not report it alone---Qwen
holds that score at a macro-F1 of $0.093$.
The specialist also leads every reasoning axis in both settings (SUM $12.57$ on
the ECG and $13.23$ on the PPG, against $9.65$ and $10.52$ for the strongest
baseline; Table~\ref{tab:spec-reason}), by the widest
margin on completeness ($3.47$ vs.\ $2.50$) and validity ($3.06$ vs.\ $2.33$), and
the narrowest on faithfulness ($2.64$ vs.\ $2.28$).
 
\begin{table}[t]
\centering\small
\caption{\textbf{ECG and PPG specialist} classification (tool level). ECG is 9-class; PPG is 7-class.}
\label{tab:spec-cls}
\setlength{\tabcolsep}{4pt}
\begin{tabular}{lccc}
\toprule
& macro-F1 & bal.\ acc & raw acc \\
\midrule
\multicolumn{4}{l}{\emph{ECG (9-class)}}\\
\textbf{Specialist (Ours)} & \textbf{0.393} & \textbf{0.423} & 0.762 \\
Gemini-2.5-Pro   & 0.131 & 0.148 & 0.314 \\
Qwen2.5-VL-72B   & 0.093 & 0.117 & 0.587 \\
MedGemma-27B     & 0.090 & 0.127 & 0.578 \\
Llama-3.2-90B-Vision    & 0.089 & 0.094 & 0.347 \\
GPT-4o           & 0.080 & 0.126 & 0.335 \\
\midrule
\multicolumn{4}{l}{\emph{PPG (7-class)}}\\
\textbf{Specialist (Ours)} & \textbf{0.367} & \textbf{0.380} & 0.783 \\
Qwen2.5-VL-72B   & 0.198 & 0.214 & 0.514 \\
Gemini-2.5-Pro   & 0.194 & 0.212 & 0.560 \\
Llama-3.2-90B-Vision    & 0.169 & 0.171 & 0.518 \\
GPT-4o           & 0.147 & 0.162 & 0.549 \\
MedGemma-27B     & 0.131 & 0.146 & 0.470 \\
\bottomrule
\end{tabular}
\end{table}
 
\begin{table}[t]
\centering\small
\caption{\textbf{ECG and PPG specialist} reasoning (tool level), LLM-judge on four
axes ($1$--$4$): correctness, faithfulness, validity, completeness; SUM$/16$.}
\label{tab:spec-reason}
\setlength{\tabcolsep}{3.5pt}
\begin{tabular}{lccccc}
\toprule
& corr & faith & valid & compl & SUM \\
\midrule
\multicolumn{6}{l}{\emph{ECG (9-class)}}\\
\textbf{Specialist (Ours)} & \textbf{3.40} & \textbf{2.64} & \textbf{3.06} & \textbf{3.47} & \textbf{12.57} \\
Qwen2.5-VL-72B   & 2.72 & 2.28 & 2.33 & 2.32 & 9.65 \\
MedGemma-27B     & 2.65 & 1.88 & 2.18 & 2.34 & 9.06 \\
GPT-4o           & 2.01 & 1.97 & 1.88 & 2.50 & 8.35 \\
Gemini-2.5-Pro   & 1.97 & 1.67 & 1.82 & 2.08 & 7.53 \\
Llama-3.2-90B-Vision    & 1.99 & 1.56 & 1.70 & 1.87 & 7.11 \\
\midrule
\multicolumn{6}{l}{\emph{PPG (7-class)}}\\
\textbf{Specialist (Ours)} & \textbf{3.35} & \textbf{3.25} & \textbf{3.25} & \textbf{3.38} & \textbf{13.23} \\
GPT-4o           & 2.65 & 2.64 & 2.60 & 2.63 & 10.52 \\
Gemini-2.5-Pro   & 2.64 & 2.57 & 2.53 & 2.61 & 10.35 \\
Qwen2.5-VL-72B   & 2.54 & 2.70 & 2.52 & 2.50 & 10.26 \\
Llama-3.2-90B-Vision    & 2.53 & 2.66 & 2.48 & 2.46 & 10.12 \\
MedGemma-27B     & 2.61 & 2.28 & 2.29 & 2.30 & 9.48 \\
\bottomrule
\end{tabular}
\end{table}
 
\section{Agent Details}
 
\paragraph{Deterministic tools.}
The analysts are deterministic tools computed from the raw signal with NeuroKit2
and SciPy, and their role is to re-examine the specialist's reading: each returns
a measured quantity that the class and evidence the specialist proposed must be
checked against, so the language model argues over tool outputs rather than the
waveform. Three tools form a shared backbone across both signals---peak
detection, rhythm analysis over the peak-to-peak series, and an artifact/quality
gate---and the rest are modality-specific, which is where sensor physics enters.
The ECG agent has seven tools in all, including those the pulse cannot reproduce:
\begin{itemize}
\item \textbf{QRS morphology} --- beat width and shape, separating wide
ventricular beats from narrow ones.
\item \textbf{Atrial activity} --- P-wave presence and organization, separating
AF from MAT and SVT.
\item \textbf{AV conduction} --- pause structure, separating sinus-node
dysfunction from AV block.
\end{itemize}
The PPG agent has nine, weighted toward what the pulse does carry:
\begin{itemize}
\item \textbf{Perfusion} --- perfusion index, for the haemodynamic significance
of a fast rhythm.
\item \textbf{Pulse-wave / beat morphology} --- pulse-wave shape and beat matching
against a template, for ectopy and aberrant beats.
\end{itemize}
One of these tools localizes the abnormal span independently of the class, for
grounding.
 
\paragraph{Within-agent debate.}
Inside each agent the window is read by the same analyst--devil's-advocate debate
used for window fusion (main text), with two differences. It has no fixed
referee: three analysts issue independent verdicts and revise over two rounds to a
majority position~\citep{du2024debate}. And it is gated by a critic---if the
devil's advocate's objection is left unresolved, the reading is failed and
returned for another round of tool use, so evidence is re-gathered rather than
assumed. The physical constraints still hold: the PPG keeps the merged VT/MAT and
PVC/PAC classes and defers to the ECG, and a PPG claim about a P wave or QRS width
is treated as a fabrication.
 
\section{Patient-Level Pipeline}
 
\paragraph{Retrieval.}
Retrieval draws on two sources: an online PubMed search through the NCBI
E-utilities, and a curated local base of about twenty guideline snippets, each a
short passage with its PMID embedded; local topics span R--R thresholds, AF on the
PPG, ectopy and PVC burden, SVT, rate ranges, and the class-specific management
references (2023 AF and CHA$_2$DS$_2$-VASc, 2017 VT and sudden death, 2018
bradycardia and AV block). Matching is TF-IDF cosine similarity over the local
base and a live query against PubMed. An LLM draws two or three queries from the
fused patient state and metadata---about significance, management, and risk, not
the diagnosis, which is already settled---and the top snippets are retrieved; the
PMID in each is surfaced as an explicit tag that later stages cite verbatim. For
the reported numbers we fix retrieval to the curated base so that every citation
resolves to a manually verified reference and the evaluation is exactly
reproducible; the PubMed path is used when the system runs live.
 
\paragraph{Critic.}
The critic checks the draft P1--P3 against a fixed rule set and edits only
violations:
\begin{itemize}
\item \textbf{Grounding} --- remove any finding not supported by the ECG or PPG
state.
\item \textbf{Citations} --- every risk or management claim must cite a PMID that
appears in the retrieved set; mismatches are re-mapped, fabricated PMIDs deleted,
missing ones added.
\item \textbf{Anticoagulation} --- not indicated for ectopy or on
CHA$_2$DS$_2$-VASc alone; a recommendation without documented AF is corrected.
\item \textbf{Redundant tests} --- the agent already holds the ECG, so any
``obtain an ECG'' line is replaced with the ECG's actual finding.
\item \textbf{Class specificity} --- a management line that would read the same
for any rhythm is rewritten to the diagnosed class.
\end{itemize}
If the critic's own output is malformed, the draft is kept.
 
\paragraph{Patient-level debate.}
The three patient stages reuse the same debate primitive, differing only in what
they weigh. Aggregation runs analysts to a referee \emph{without} a devil's
advocate and is descriptive: one lens tracks the dominant rhythm and burden, the
other enumerates every distinct arrhythmia without ranking, and the PPG records
its own AF confidence without asserting an atrial rhythm it cannot confirm.
Fusion and reasoning both add a devil's advocate. Fusion reconciles the two
states---ECG for QRS and P-waves, PPG for pulse regularity and perfusion---and
then applies severity, its referee suppressing PPG-only AF false positives and
taking the dominant rhythm as the one the ECG shows across most windows. Reasoning
adds the metadata to produce P1--P3, its devil attacking mis-prioritized claims
(benign ectopy raised over a present AF or VT, urgency not matching the evidence)
and its referee requiring that P2 and P3 cite only retrieved PMIDs and that P3 be
class-specific.
 
\section{Scoring Rubrics}
 
\paragraph{Why anchored rubrics.} We score reasoning with an LLM judge because the
quality we care about---whether a reading is correct, grounded in real evidence,
and clinically actionable---has no reference string to match and is exactly what
reference-free LLM evaluation was introduced to
capture~\citep{zheng2023judge,liu2023geval,gu2024judgesurvey}. A free-form quality
score from such a judge is unstable and drifts between calls, so we do not ask for
one: following the form-filling paradigm of G-Eval~\citep{liu2023geval}, each axis
is decomposed into explicit per-grade anchors, and the judge selects the grade
whose description the output meets rather than inventing a number. Anchoring the
grades this way is what makes the scores reproducible and auditable---the reason a
given output received a given grade is written into the rubric---and it constrains
the judge to the clinical criteria below rather than its own priors. To remove the
self-preference bias documented for LLM
judges~\citep{liu2023geval,zheng2023judge}, the judge (DeepSeek-V4-Pro) is drawn
from outside the baseline set, so no system grades its own outputs.
 
\paragraph{Where the anchors come from.} The main-text Metrics section defines the
four reasoning axes and the T1--T3 and P1--P3 tasks; here we give the per-grade
anchors the judge applies. The four window axes ($4$/$3$/$2$/$1$) are shared
across the specialist reading and the window probes, specialized by task (T1
diagnosis, T2 differential, T3 counterfactual) and by setting (both: nine-class;
PPG: seven-class merged). The patient anchors ($4$/$3$/$2$/$1$/$0$, adding a
\emph{$0$} for output that does not address the axis at all) are written against
clinical practice: acuity follows ACLS triage, and management and significance
follow the 2023 AF~/~CHA$_2$DS$_2$-VASc~\citep{joglar2024af}, 2017
VT/SCD~\citep{alkhatib2018va}, and 2018 bradycardia~\citep{kusumoto2019brady}
guidelines, so that a
top-grade answer is not merely fluent but concordant with the standard of care.
The PPG anchors additionally encode signal physics---a claim to a P~wave or a QRS
width from the pulse is capped as unfaithful, because the pulse cannot show
either.

\begin{rubricbox}{Window-level rubric: four axes, shared by the specialist and T1/T2/T3 (grades $4{\to}1$)}
\small
\rubtask{T1 --- diagnosis}
\rubaxis{Correctness}
{exact GT class.}
{adjacent family (PVC$\leftrightarrow$PAC, VT$\leftrightarrow$MAT, AVB$\leftrightarrow$SND).}
{right normal-vs-arrhythmia side, wrong family.}
{wrong side or nonsensical.}
\rubaxis{Faithfulness}
{engages measured rate / R--R~CV / QRS, numbers within ${\sim}25\%$, no hallucination.}
{one minor issue (a number off $25$--$50\%$).}
{a qualitative claim contradicts evidence, or a number off ${>}50\%$.}
{fabricated or $2\times$-wrong values.}
\rubaxis{Validity}
{cited features correct and sufficient.}
{valid but incomplete.}
{some features irrelevant or contradictory.}
{features do not support the conclusion.}
\rubaxis{Completeness}
{all discriminators addressed (rate, R--R regularity, QRS/ventricular morphology, pauses).}
{one missing.}
{two or more missing.}
{conclusory, no feature support.}
 
\rubtask{T2 --- differential}
\rubaxis{Correctness}
{correct primary and the key alternatives, incl.\ the most dangerous.}
{correct primary, differential incomplete.}
{primary a near-miss, or a weak set.}
{wrong primary or irrelevant differentials.}
\rubaxis{Faithfulness}
{every exclusion cites a real evidence-based reason.}
{one exclusion weakly grounded.}
{an exclusion contradicts evidence or is empty.}
{exclusions largely fabricated.}
\rubaxis{Validity}
{each alternative ruled out by a correct discriminator.}
{one weak discriminator.}
{some exclusions illogical.}
{logic broken or circular.}
\rubaxis{Completeness}
{covers the relevant alternatives and explicitly the most dangerous.}
{one notable alternative missing.}
{omits the most dangerous.}
{no real differential.}
 
\rubtask{T3 --- counterfactual}
\rubaxis{Correctness}
{correct pivotal change and resulting diagnosis.}
{correct change but adjacent result (or vice versa).}
{a plausible change, but not the gold one.}
{wrong or irrelevant.}
\rubaxis{Faithfulness}
{the change is evidence-relevant and realistic.}
{minor imprecision.}
{contradicts current evidence, or vague.}
{fabricated or irrelevant.}
\rubaxis{Validity}
{causal link textbook-correct.}
{plausible but loose.}
{weak or partial link.}
{change does not lead to the result.}
\rubaxis{Completeness}
{both the change and the result explicit.}
{one slightly vague.}
{only one given.}
{no real counterfactual.}
\end{rubricbox}
 
\paragraph{PPG (seven-class) specialization.}
Under PPG, correctness is satisfied by naming a ``premature/ectopic beat'' ($\to$PVC/PAC) or a ``broad/fast run'' ($\to$VT/MAT) and keeping an ECG-only-distinguishable pair merged, deferring to the ECG; excluding an indistinguishable alternative is penalized. Faithfulness caps at $2$ for any claim to see P-waves or QRS, which the pulse cannot show, and completeness rewards flagging that an ECG is required where atrial or ventricular detail is needed.
 
\begin{rubricbox}{Patient-level rubric: P1/P2/P3, two axes each (grades $4{\to}0$; the two axes sum to $8$)}
\small
\rubtask{P1 --- impression}
\rubaxisF{Correctness --- were the arrhythmias present named?}
{named every arrhythmia present and asserted none that were absent.}
{named the present arrhythmias but missed one, or added an absent one as a secondary mention.}
{named a present arrhythmia only as a secondary mention, and asserted an absent one as dominant.}
{mentioned a present arrhythmia at least once.}
{named none of the arrhythmias present.}
\rubaxisF{Faithfulness --- do the cited features fit the named arrhythmia?}
{cited only features characteristic of the named arrhythmia (e.g.\ AF~$\to$~irregularly irregular R--R).}
{cited features of both the present and an absent arrhythmia.}
{cited a feature of an absent arrhythmia, though the feature itself is a valid basis for that arrhythmia (right features, wrong diagnosis).}
{cited a feature that supports no arrhythmia, or one physically impossible (a P-wave/PR interval on PPG, R--R~CV~$0.06$).}
{cited no features (diagnosis named only).}
 
\rubtask{P2 --- clinical significance}
\rubaxisF{Prioritization --- is the most dangerous finding foregrounded? (severity, not frequency)}
{foregrounded the most dangerous finding as the one to act on, with severity stated correctly.}
{correct priority, but severity over- or under-stated.}
{mentioned the dangerous finding but did not prioritize it, or prioritized a less dangerous one.}
{did not mention the most dangerous finding; presented only a less dangerous one as the one to act on.}
{prioritized no finding at all.}
\rubaxisF{Context --- is the risk tied to why this patient is at risk? (e.g.\ 70~yr, hypertension~$\to$~AF~$\to$~stroke)}
{``this patient is at risk \emph{because}...'' holds---patient features connect to a specific risk.}
{the link holds but omits one relevant feature (e.g.\ names AF risk without the comorbidity driving it).}
{patient data restated but not tied to risk---true of any patient.}
{used a feature unrelated to the arrhythmia as the risk basis, or judged risk backwards.}
{used no patient information.}
 
\rubtask{P3 --- management}
\rubaxisF{Management --- is the plan specific to this rhythm and this patient?}
{specific to the rhythm and patient, guideline-concordant, with a correct citation.}
{appropriate but generic, or the citation is loose.}
{a generic plan that would read the same for any rhythm (no rhythm-specific decision).}
{an inappropriate plan (anticoagulation without documented AF, discharging a sustained VT).}
{gave no management plan.}
\rubaxisF{Urgency \& safety --- is the urgency right and free of over-/under-treatment?}
{urgency correct and management connects to real findings safely.}
{urgency slightly off but still safe.}
{urgency clearly wrong, or a generic ``urgent'' label not tied to a specific finding.}
{harmful (discharges a VT, wrong anticoagulation, a life-threatening rhythm missed).}
{did not address urgency at all.}
\end{rubricbox}
 
\section{Baseline Prompts}
 
Every baseline is given the same information the task provides for answering---the
raw ECG and PPG, their plots, the taxonomy, and, at the patient level, the
metadata and every window---and is asked for identical outputs. What our system
adds is not more information but its orchestration over that shared input, so the
comparison isolates the agent construction. At the window
level a baseline receives both the ECG and the PPG---their raw sample values and
their plots---together with the taxonomy, and is asked for the class, the span,
and T1/T2/T3 in one call. The difference from our agent is the tool layer it does
not receive: the extracted measurements, R-peak coordinates, morphological hints,
and the clinical-reasoning scaffold. All baseline calls use temperature~$0$ and
request a single JSON object.
 
\begin{promptbox}{Baseline prompt --- window level (both signals, nine-class)}
\small
\prole{System.} \textit{``You are an expert cardiologist. Analyze the signal(s)
and image(s) and answer ALL items.''}
 
\prole{User message.} A multimodal message interleaving two images and text:
\begin{quote}\small\itshape
\upshape\texttt{[ECG plot image]}\quad\texttt{[PPG plot image]}\itshape\\[3pt]
You are given two time-aligned signals of the same cardiac window. The first image
is the ECG, the second is the PPG. ECG signal values (index $0\dots N{-}1$):
$87, 42, 11, 17, 3, 9, \dots$ (integer, ${\times}100$). PPG signal values:
$88, 22, 14, 28, 8, 11, \dots$ Then the nine-class arrhythmia reference, each class
with its criteria (AF: irregularly irregular R--R, CV${>}0.15$, absent P waves,
narrow QRS; VT: fast run of wide QRS; and so on for SVT, PVC, PAC, SND, MAT, AVB,
Normal).
\end{quote}
 
\prole{Requested output.} A single JSON object
\texttt{\{"class","span","T1","T2","T3"\}}, where:
\begin{itemize}\itemsep1pt
\item \texttt{class}: the single best-fitting class from the taxonomy;
\texttt{span}: the sample range \texttt{[start,end]} of the abnormal segment.
\item \texttt{T1}: diagnose the rhythm and the key features (rate, R--R regularity,
P-wave, QRS morphology) that support it.
\item \texttt{T2}: the plausible differentials and why each is excluded, the most
dangerous first.
\item \texttt{T3}: the single change that would most alter the diagnosis, and to
what.
\end{itemize}
\prole{Decoding.} \texttt{temperature}${=}0$, \texttt{max\_tokens}${=}900$,
\texttt{response\_format}${=}$\texttt{json\_object}.
 
\smallskip\noindent
Unlike our specialist, the baseline is given no extracted features, R-peak
coordinates, morphological hints, or reasoning scaffold; it reads the class
directly from the raw signal and plots.
\end{promptbox}
 
\begin{promptbox}{Baseline prompt --- patient level (P1/P2/P3)}
\small
\prole{System.} \textit{``You are a cardiologist reviewing a patient's
peri-operative continuous cardiac monitoring. For the same patient you are given
several time windows; for each window you receive the ECG and PPG raw waveforms
($100$~Hz, ${\sim}10$~s) and their plots, plus clinical metadata. Produce an
integrated patient-level assessment. Ground every statement in the actual
waveform/plot findings and the metadata---do not invent findings that are not
supported by the data.''}
 
\prole{User message.} The patient's metadata followed by every window in temporal
order, laid out as:
\begin{quote}\small
\itshape PATIENT METADATA: Age${=}78$; Sex${=}$female; BMI${=}25.0$;
Hypertension${=}$yes; Diabetes${=}$no. This patient has 5 monitoring windows, in
temporal order below.\\[3pt]
\upshape\texttt{=== Window 1 ===}\quad\itshape ECG raw ($100$\,Hz): [\dots]; PPG
raw ($100$\,Hz): [\dots]; \upshape\texttt{[ECG plot image]}\
\texttt{[PPG plot image]}\\[2pt]
\upshape\texttt{=== Window 2 ===}\ \itshape\dots\ \upshape(through window 5)
\end{quote}
Every window thus supplies four inputs---its ECG and PPG raw waveforms and its two
plot images---and all of the patient's windows are included.
 
\prole{Requested output.} Compact JSON \texttt{\{"P1","P2","P3"\}}, grounded only
in the provided windows and metadata:
\begin{itemize}\itemsep1pt
\item \texttt{P1} (integrated diagnosis): across all windows, which arrhythmia(s)
appear in which window(s), and the dominant / most representative rhythm, citing
specific waveform features.
\item \texttt{P2} (contextual significance): given the metadata (age, surgery,
comorbidities), the clinical significance and the single finding most important
to act on, and why.
\item \texttt{P3} (urgency \& management): the appropriate response (further
monitoring vs.\ intervention, and urgency level) and the specific evidence
supporting it.
\end{itemize}
\end{promptbox}
 
\begin{table}[!ht]
\centering\footnotesize
\setlength{\tabcolsep}{3.2pt}
\caption{Per-class F1. Setting \emph{both} is nine-class (ECG specialist);
\emph{PPG} is seven-class merged (PPG specialist).}
\label{tab:percls}
\begin{tabular}{@{}lccccccc@{}}
\toprule
class & spec. & \textbf{\textsc{Cardio.}} & gpt-4o & gem. & qwen & medg. & llama \\
\midrule
\multicolumn{8}{@{}l}{\emph{Setting: both (nine-class)}}\\
normal          & 0.93 & \textbf{0.95} & 0.84 & 0.42 & 0.77 & 0.75 & 0.00 \\
atrial fibrill. & 0.68 & \textbf{0.83} & 0.65 & 0.42 & 0.05 & 0.00 & 0.18 \\
MAT             & 0.42 & \textbf{0.43} & 0.01 & 0.00 & 0.00 & 0.00 & 0.00 \\
SVT             & \textbf{0.11} & 0.08 & 0.04 & 0.04 & 0.00 & 0.00 & 0.00 \\
VT              & 0.00 & 0.00 & 0.25 & 0.25 & 0.00 & 0.00 & 0.00 \\
PVC             & 0.50 & \textbf{0.61} & 0.21 & 0.48 & 0.43 & 0.00 & 0.00 \\
PAC             & 0.17 & \textbf{0.45} & 0.30 & 0.04 & 0.20 & 0.00 & 0.00 \\
AVB             & \textbf{0.18} & 0.12 & 0.00 & 0.00 & 0.00 & 0.00 & 0.00 \\
SND             & 0.55 & \textbf{0.57} & 0.00 & 0.02 & 0.00 & 0.00 & 0.00 \\
macro-F1        & 0.39 & \textbf{0.45} & 0.26 & 0.19 & 0.16 & 0.08 & 0.02 \\
\midrule
\multicolumn{8}{@{}l}{\emph{Setting: PPG (seven-class merged)}}\\
normal          & 0.94 & \textbf{0.93} & 0.72 & 0.70 & 0.67 & 0.66 & 0.70 \\
atrial fibrill. & \textbf{0.76} & 0.75 & 0.30 & 0.48 & 0.38 & 0.18 & 0.26 \\
VT/MAT          & 0.22 & \textbf{0.38} & 0.00 & 0.00 & 0.00 & 0.00 & 0.04 \\
SVT             & 0.00 & \textbf{0.12} & 0.00 & 0.00 & 0.00 & 0.03 & 0.06 \\
PVC/PAC         & \textbf{0.53} & 0.51 & 0.01 & 0.18 & 0.33 & 0.05 & 0.13 \\
AVB             & 0.00 & \textbf{0.04} & 0.00 & 0.00 & 0.00 & 0.00 & 0.00 \\
SND             & 0.13 & \textbf{0.24} & 0.00 & 0.00 & 0.00 & 0.00 & 0.00 \\
macro-F1        & 0.37 & \textbf{0.43} & 0.15 & 0.19 & 0.20 & 0.13 & 0.17 \\
\bottomrule
\end{tabular}
\end{table}
 
\begin{figure}[!t]
\centering
\setlength{\tabcolsep}{3pt}
\begin{tabular}{@{}r@{\hspace{4pt}}c@{}}
\footnotesize GT                     & \includegraphics[width=6cm]{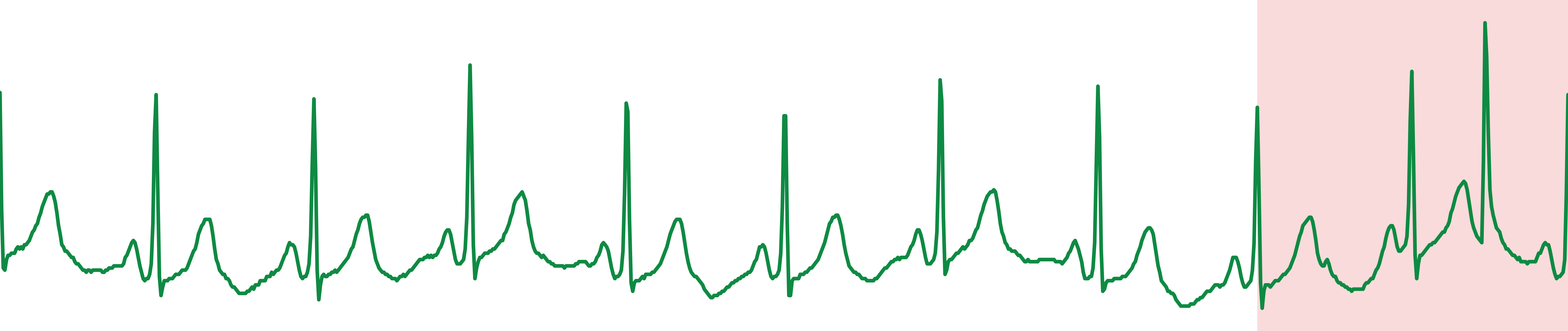} \\[2pt]
\footnotesize \textbf{Cardiologent}  & \includegraphics[width=6cm]{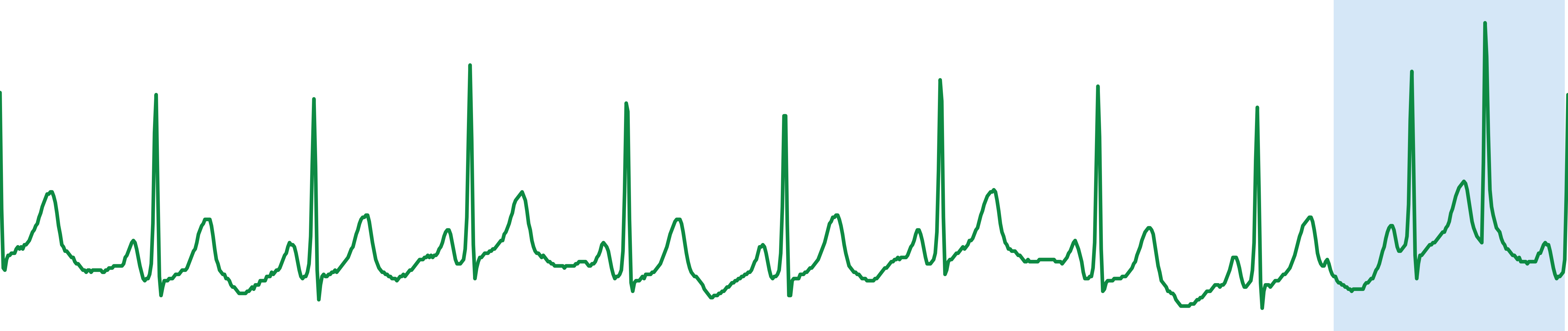} \\[2pt]
\footnotesize GPT-4o                 & \includegraphics[width=6cm]{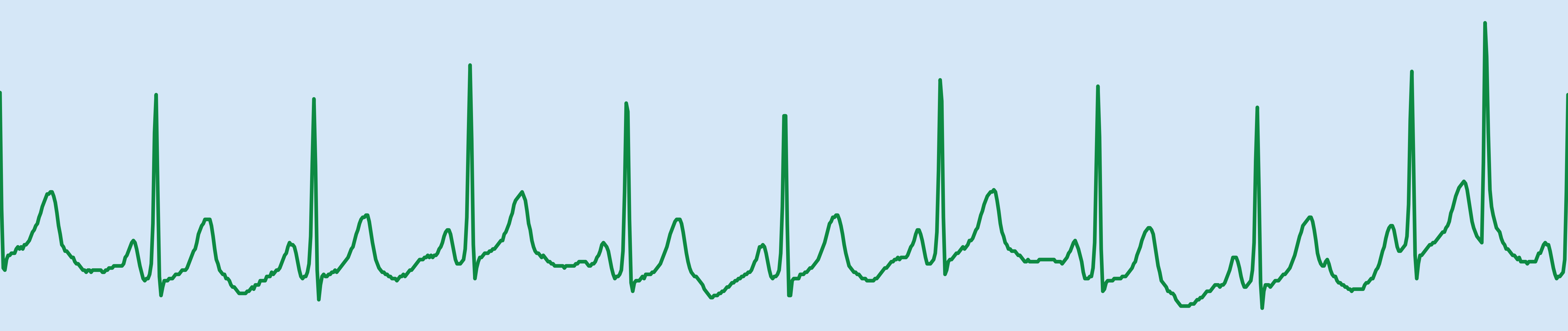} \\[2pt]
\footnotesize Gemini-2.5-Pro         & \includegraphics[width=6cm]{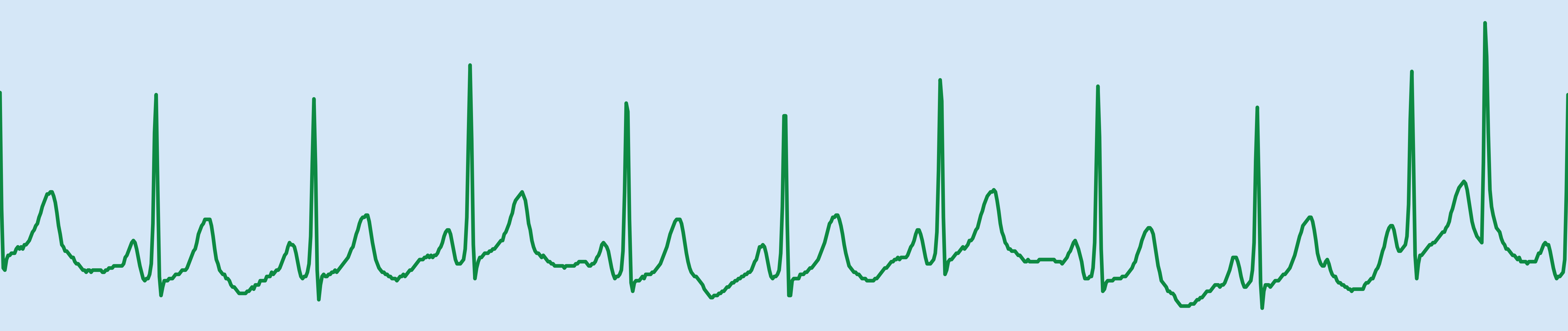} \\[2pt]
\footnotesize Qwen2.5-VL             & \includegraphics[width=6cm]{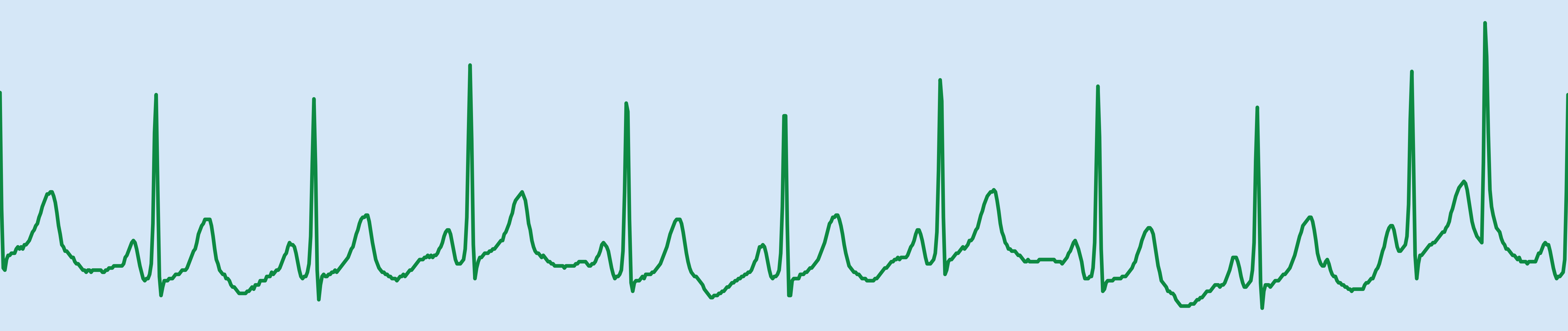} \\[2pt]
\footnotesize MedGemma               & \includegraphics[width=6cm]{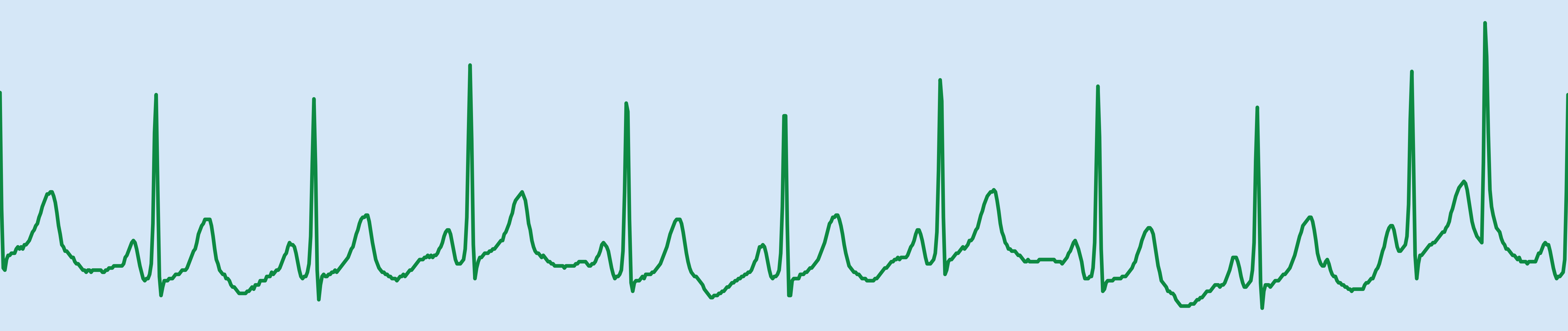} \\[2pt]
\footnotesize Llama-3.2-Vision            & \includegraphics[width=6cm]{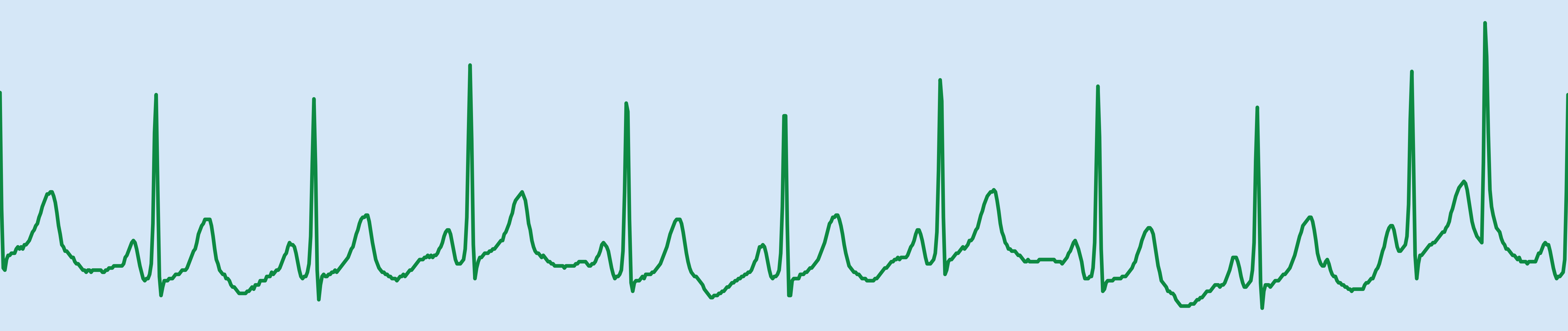} \\
\end{tabular}
\caption{Grounding on ECG window \texttt{4925\_w0099} (ground-truth PAC). Each row
is one model's localized abnormal span on the same trace; the top row is the
ground truth. \textsc{Cardiologent} localizes a tight span on the PAC, whereas the
baselines mark almost the entire window, inflating IoU while collapsing MCC toward
zero.}
\label{fig:grounding}
\end{figure}
 
\section{Experiments Details}
 
\begin{figure}[!ht]
\centering
\includegraphics[width=\linewidth]{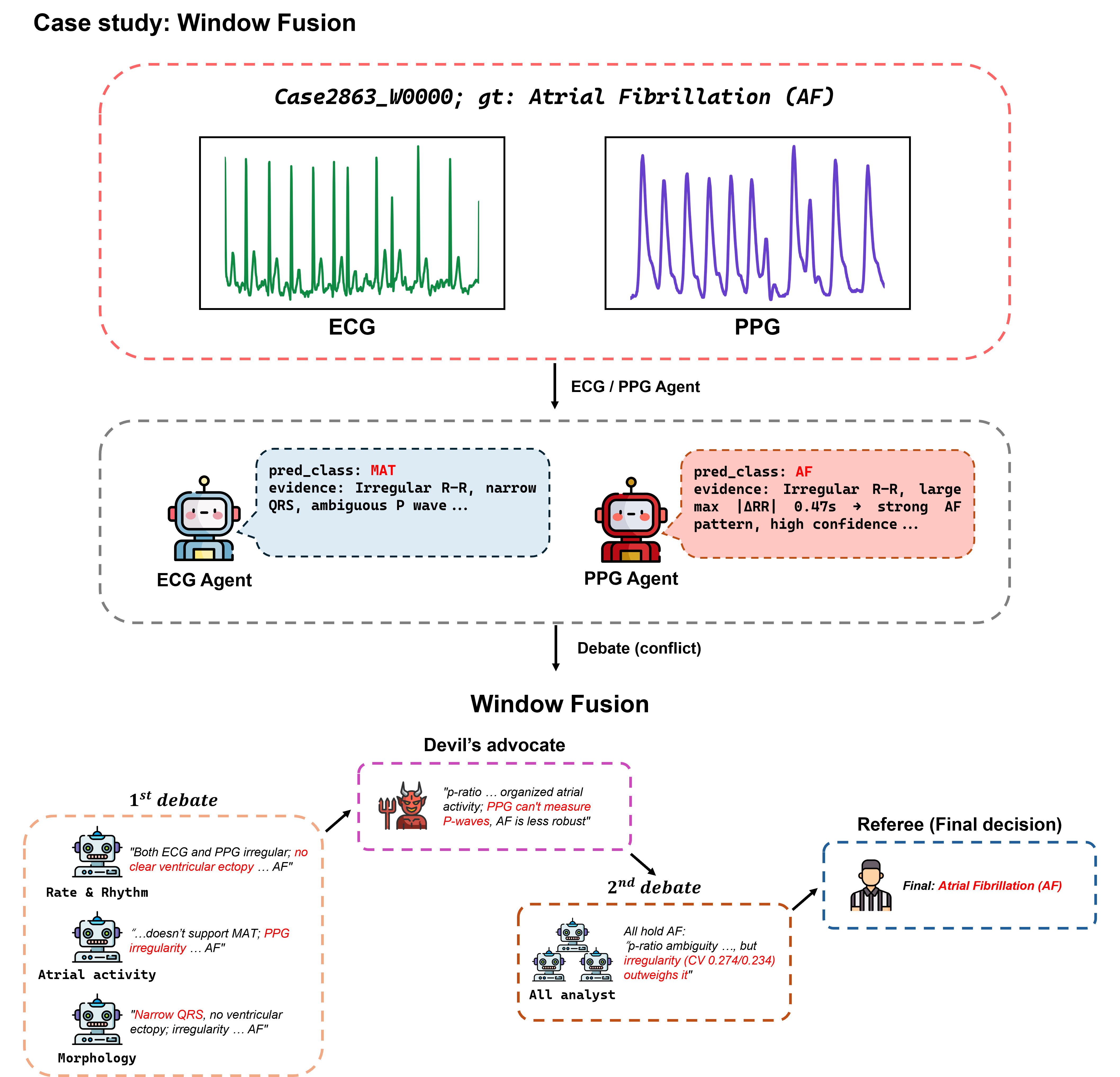}
\caption{\textbf{Case study: Window fusion} (Case2863\_W0000; GT: atrial
fibrillation). The ECG agent misreads an ambiguous P~wave and calls MAT; the PPG
agent calls AF from the R--R irregularity. Because the two conflict, fusion runs a
two-round analyst--devil's-advocate debate: the devil's advocate notes the PPG
cannot confirm P~waves, the analysts weigh this against the measured irregularity
(R--R CV $0.274$/$0.234$), and the referee settles on the correct class, AF. The
right answer is reached by weighing evidence, not by deferring to the more
confident agent.}
\label{fig:casestudy}
\end{figure}

\paragraph{Per-class classification.}
Table~\ref{tab:percls} breaks the window-level macro-F1 out by class, and it
locates where our margin comes from. It is not a uniform edge on the common
rhythms: it is that the general baselines \emph{collapse to zero} on the rare
arrhythmias. On \emph{both} signals, every baseline scores $0.00$--$0.01$ on MAT,
SVT, AVB, and SND---four of the nine classes they effectively cannot detect at
all---while our system holds usable F1 on each ($0.43$, $0.08$, $0.12$, $0.57$).
This is what the $1.8\times$ macro-F1 gap ($0.449$ vs.\ $0.255$) is made of: not a
higher ceiling on normal and AF, where the stronger baselines are already
competent, but coverage of the rare classes where they read nothing. Within our
own stack, the agent layer helps most on the classes the specialist reads least
well---PAC more than doubles ($0.17{\to}0.45$) once the atrial-activity and
premature-beat tools weigh in, and AF ($0.68{\to}0.83$) and PVC
($0.50{\to}0.61$) improve likewise---though it is not uniformly additive: on SVT
and AVB the extra deliberation slightly lowers an already low score. The help is
largest exactly where the raw signal is thinnest: under \emph{PPG}, where the
specialist is weak on almost every arrhythmia, the agent lifts nearly all of
them---SVT and AVB from $0.00$ to $0.12$ and $0.04$, SND $0.13{\to}0.24$, and the
merged VT/MAT $0.22{\to}0.38$---so the tools and debate matter more, not less,
when the pulse alone carries the reading. The one class we do not recover is VT:
with $45$ test windows on \emph{both} it is too rare for the specialist to learn,
and here two baselines edge us ($0.25$ vs.\ $0.00$), though under \emph{PPG} the
merged VT/MAT class reaches $0.38$. Figure~\ref{fig:casestudy} walks through a
window where the ECG and PPG agents first disagree and window fusion settles them
on the correct class.

\begin{table*}[!ht]
\centering\footnotesize
\setlength{\tabcolsep}{3pt}
\caption{Reasoning by rubric axis (LLM-judge, $1$--$4$). Window probes have four
axes (SUM$/16$); patient tasks two axes (SUM$/8$). Best per column in bold.}
\label{tab:reason-axis}
\begin{tabular}{@{}l ccccc ccccc ccccc@{}}
\toprule
\multicolumn{16}{@{}l}{\emph{Window level --- Setting both}}\\
& \multicolumn{5}{c}{T1 (diagnosis)} & \multicolumn{5}{c}{T2 (differential)} & \multicolumn{5}{c}{T3 (counterfactual)} \\
\cmidrule(lr){2-6}\cmidrule(lr){7-11}\cmidrule(lr){12-16}
model & corr & faith & valid & compl & SUM & corr & faith & valid & compl & SUM & corr & faith & valid & compl & SUM \\
\midrule
\textsc{Cardiologent} & \textbf{3.54} & \textbf{2.72} & \textbf{3.22} & \textbf{3.03} & \textbf{12.51} & \textbf{2.34} & \textbf{2.60} & 2.81 & \textbf{2.75} & \textbf{10.50} & \textbf{2.44} & \textbf{3.47} & \textbf{3.21} & \textbf{3.76} & \textbf{12.88} \\
gpt-4o & 3.17 & 2.55 & 2.92 & 2.94 & 11.57 & 1.65 & 2.40 & 2.69 & 2.25 & 8.99 & 2.34 & 3.37 & 2.79 & 3.53 & 12.02 \\
gemini & 2.24 & 1.59 & 1.99 & 2.84 & 8.66 & 2.14 & 2.52 & \textbf{2.84} & 2.44 & 9.94 & 2.02 & 2.98 & 2.83 & 3.48 & 11.30 \\
qwen & 2.86 & 2.32 & 2.73 & 2.83 & 10.74 & 1.13 & 1.32 & 1.54 & 1.55 & 5.55 & 1.41 & 2.05 & 2.15 & 2.61 & 8.21 \\
medgemma & 2.83 & 1.71 & 2.46 & 2.75 & 9.75 & 1.02 & 1.13 & 1.33 & 1.39 & 4.87 & 1.24 & 1.95 & 2.12 & 1.90 & 7.21 \\
llama & 1.12 & 1.05 & 1.09 & 1.20 & 4.47 & 1.06 & 1.22 & 1.30 & 1.20 & 4.78 & 1.30 & 2.26 & 1.66 & 2.36 & 7.58 \\
\midrule
\multicolumn{16}{@{}l}{\emph{Window level --- Setting PPG}}\\
& \multicolumn{5}{c}{T1 (diagnosis)} & \multicolumn{5}{c}{T2 (differential)} & \multicolumn{5}{c}{T3 (counterfactual)} \\
\cmidrule(lr){2-6}\cmidrule(lr){7-11}\cmidrule(lr){12-16}
model & corr & faith & valid & compl & SUM & corr & faith & valid & compl & SUM & corr & faith & valid & compl & SUM \\
\midrule
\textsc{Cardiologent} & \textbf{3.32} & \textbf{3.48} & \textbf{3.35} & 2.74 & \textbf{12.90} & \textbf{1.96} & 2.23 & \textbf{2.11} & \textbf{1.95} & \textbf{8.26} & \textbf{2.32} & \textbf{3.36} & \textbf{2.80} & \textbf{3.70} & \textbf{12.19} \\
gpt-4o & 2.10 & 2.13 & 2.07 & 1.90 & 8.21 & 1.16 & \textbf{2.47} & 2.07 & 1.31 & 7.01 & 2.12 & 3.03 & 2.72 & 3.43 & 11.30 \\
gemini & 2.82 & 2.32 & 2.59 & 2.77 & 10.50 & 1.30 & 1.92 & 1.93 & 1.41 & 6.56 & 1.38 & 2.37 & 1.89 & 3.64 & 9.29 \\
qwen & 2.77 & 2.71 & 2.71 & \textbf{2.85} & 11.03 & 1.09 & 2.03 & 2.05 & 1.48 & 6.65 & 1.17 & 2.28 & 1.71 & 2.47 & 7.63 \\
medgemma & 2.75 & 2.48 & 2.61 & 2.77 & 10.60 & 1.06 & 1.76 & 1.71 & 1.24 & 5.78 & 1.17 & 2.27 & 1.74 & 2.04 & 7.22 \\
llama & 1.70 & 1.35 & 1.41 & 1.33 & 5.80 & 1.58 & 2.05 & 1.92 & 1.53 & 7.07 & 1.56 & 2.37 & 1.78 & 2.67 & 8.38 \\
\midrule
\multicolumn{16}{@{}l}{\emph{Patient level -- judge (65 patients)}}\\
& \multicolumn{5}{c}{P1 (diagnosis)} & \multicolumn{5}{c}{P2 (significance)} & \multicolumn{5}{c}{P3 (management)} \\
\cmidrule(lr){2-6}\cmidrule(lr){7-11}\cmidrule(lr){12-16}
model & \multicolumn{2}{c}{correct} & \multicolumn{2}{c}{faithful} & SUM & \multicolumn{2}{c}{priority} & \multicolumn{2}{c}{context} & SUM & \multicolumn{2}{c}{mgmt} & \multicolumn{2}{c}{urgency} & SUM \\
\midrule
\textsc{Cardiologent} & \multicolumn{2}{c}{\textbf{1.97}} & \multicolumn{2}{c}{\textbf{2.69}} & \textbf{4.66} & \multicolumn{2}{c}{\textbf{2.34}} & \multicolumn{2}{c}{\textbf{2.03}} & \textbf{4.37} & \multicolumn{2}{c}{\textbf{2.14}} & \multicolumn{2}{c}{\textbf{2.42}} & \textbf{4.55} \\
gemini & \multicolumn{2}{c}{0.89} & \multicolumn{2}{c}{2.31} & 3.20 & \multicolumn{2}{c}{1.69} & \multicolumn{2}{c}{1.62} & 3.31 & \multicolumn{2}{c}{1.85} & \multicolumn{2}{c}{2.09} & 3.94 \\
gpt-4o & \multicolumn{2}{c}{0.55} & \multicolumn{2}{c}{2.23} & 2.78 & \multicolumn{2}{c}{1.48} & \multicolumn{2}{c}{1.35} & 2.83 & \multicolumn{2}{c}{1.37} & \multicolumn{2}{c}{1.57} & 2.94 \\
qwen & \multicolumn{2}{c}{0.12} & \multicolumn{2}{c}{2.00} & 2.12 & \multicolumn{2}{c}{1.06} & \multicolumn{2}{c}{1.17} & 2.23 & \multicolumn{2}{c}{1.29} & \multicolumn{2}{c}{1.57} & 2.86 \\
medgemma & \multicolumn{2}{c}{0.00} & \multicolumn{2}{c}{1.91} & 1.91 & \multicolumn{2}{c}{0.85} & \multicolumn{2}{c}{1.25} & 2.09 & \multicolumn{2}{c}{1.32} & \multicolumn{2}{c}{1.57} & 2.89 \\
llama & \multicolumn{2}{c}{0.12} & \multicolumn{2}{c}{1.94} & 2.06 & \multicolumn{2}{c}{0.88} & \multicolumn{2}{c}{0.62} & 1.49 & \multicolumn{2}{c}{1.38} & \multicolumn{2}{c}{1.52} & 2.91 \\
\midrule
\multicolumn{16}{@{}l}{\emph{Patient level --- cardiologists (20 patients)}}\\
& \multicolumn{5}{c}{P1 (diagnosis)} & \multicolumn{5}{c}{P2 (significance)} & \multicolumn{5}{c}{P3 (management)} \\
\cmidrule(lr){2-6}\cmidrule(lr){7-11}\cmidrule(lr){12-16}
model & \multicolumn{2}{c}{correct} & \multicolumn{2}{c}{faithful} & SUM & \multicolumn{2}{c}{priority} & \multicolumn{2}{c}{context} & SUM & \multicolumn{2}{c}{mgmt} & \multicolumn{2}{c}{urgency} & SUM \\
\midrule
\textsc{Cardiologent} & \multicolumn{2}{c}{\textbf{2.23}} & \multicolumn{2}{c}{\textbf{2.35}} & \textbf{4.58} & \multicolumn{2}{c}{\textbf{2.35}} & \multicolumn{2}{c}{\textbf{2.35}} & \textbf{4.70} & \multicolumn{2}{c}{\textbf{2.52}} & \multicolumn{2}{c}{\textbf{2.50}} & \textbf{5.03} \\
gemini & \multicolumn{2}{c}{1.62} & \multicolumn{2}{c}{1.70} & 3.33 & \multicolumn{2}{c}{1.82} & \multicolumn{2}{c}{1.93} & 3.75 & \multicolumn{2}{c}{2.08} & \multicolumn{2}{c}{2.15} & 4.22 \\
gpt-4o & \multicolumn{2}{c}{1.40} & \multicolumn{2}{c}{1.57} & 2.98 & \multicolumn{2}{c}{1.65} & \multicolumn{2}{c}{1.82} & 3.48 & \multicolumn{2}{c}{1.88} & \multicolumn{2}{c}{1.98} & 3.85 \\
qwen & \multicolumn{2}{c}{0.68} & \multicolumn{2}{c}{0.80} & 1.48 & \multicolumn{2}{c}{1.18} & \multicolumn{2}{c}{1.30} & 2.48 & \multicolumn{2}{c}{1.43} & \multicolumn{2}{c}{1.50} & 2.92 \\
medgemma & \multicolumn{2}{c}{0.55} & \multicolumn{2}{c}{0.82} & 1.38 & \multicolumn{2}{c}{1.30} & \multicolumn{2}{c}{1.32} & 2.62 & \multicolumn{2}{c}{1.27} & \multicolumn{2}{c}{1.52} & 2.80 \\
llama & \multicolumn{2}{c}{0.40} & \multicolumn{2}{c}{0.60} & 1.00 & \multicolumn{2}{c}{0.88} & \multicolumn{2}{c}{1.00} & 1.88 & \multicolumn{2}{c}{1.05} & \multicolumn{2}{c}{1.20} & 2.25 \\
\bottomrule
\end{tabular}
\end{table*}
 
\paragraph{Grounding.}
Figure~\ref{fig:grounding} shows where each model localizes the abnormality on
ECG window \texttt{4925\_w0099}, whose ground truth is a premature atrial
contraction (PAC). A PAC is a focal event confined to a few beats, so a correct
localization is a narrow span. \textsc{Cardiologent} places a tight span on the
PAC, while the general baselines mark essentially the whole window as
abnormal---failing to localize. Marking everything inflates the overlap-based IoU
yet carries no localizing information, which is exactly what the per-sample MCC
exposes: a whole-window prediction overlaps the truth by chance (high IoU) but
falls to $\mathrm{MCC}\approx 0$. The split shows in the window-level grounding results: Llama posts a competitive IoU ($0.533$, second only to
ours on \emph{both}) but an MCC of $0.001$---overlap without localization---and
across all baselines none exceeds MCC $0.18$, with MedGemma going negative, even
though several reach moderate IoU. \textsc{Cardiologent} leads on both
($0.584$~/~$0.305$). We therefore report both. Let $P$ be the set of samples a
model predicts abnormal and $G$ the true abnormal samples:
\begin{align*}
\mathrm{IoU} &= \frac{|P \cap G|}{|P \cup G|}, \\[4pt]
\mathrm{MCC} &= \frac{TP\cdot TN - FP\cdot FN}
{\sqrt{(TP{+}FP)(TP{+}FN)(TN{+}FP)(TN{+}FN)}}.
\end{align*}
IoU rewards overlap, while MCC scores per-sample agreement and cannot be inflated
by a blanket abnormal prediction (which drives $TN$ to zero).

\paragraph{Reasoning by axis.}
Table~\ref{tab:reason-axis} gives every rubric axis behind the reasoning scores,
with the per-probe SUM (window SUM over four axes, max $16$; patient SUM over two
axes, max $8$). Read by axis rather than by SUM, it shows \emph{where} the margin
lives and where it does not.
 
\emph{Window level.} Our advantage is a reading whose four axes hold
\emph{together}: the cited features are real (faithful), they actually support the
diagnosis (valid), and the discriminators are covered (complete), not just a
correct label. This shows in how the axes move as a block rather than one at a
time. On T1 our SUM leads ($12.51$ vs.\ GPT-4o's
$11.57$ on \emph{both}), and the widest single-axis gaps are in validity and
completeness ($3.22$ vs.\ Gemini's $1.99$; $3.03$ vs.\ $2.84$)---the axes that ask
whether the evidence is wired to the conclusion, which is what the tools and the
debate enforce. A baseline that lands the right label often does not carry the rest
of the reading with it: Gemini's T1 correctness is $2.24$ but its faithfulness
falls to $1.59$, an answer that is right without being grounded. T2---the
differential---is the hardest probe for everyone (correctness the lowest cell for
every model, our $2.34$ vs.\ the baselines' $1.02$--$2.14$), since ruling
alternatives out demands more than naming the rhythm, and it is where completeness
most separates us ($2.75$ vs.\ Gemini's $2.44$). T3 (counterfactual) is where the
stronger baselines come closest, as reasoning about a hypothetical change draws on
general-purpose ability rather than on reading the trace ($12.02$ vs.\ our $12.88$
on \emph{both}). The few cells a baseline tops are single axes of a single
probe---Gemini's T2 validity on \emph{both} ($2.84$ vs.\ $2.81$), GPT-4o's T2
faithfulness on \emph{PPG}---and none carries the probe it sits in. The consistent
read is that our reasoning is not merely more often right but more often
\emph{coherent}: the diagnosis, its evidence, and its differential agree, which is
exactly the property a downstream patient-level decision has to rest on.
 
\emph{Patient level.} Here the axis split is starkest, and it exposes a specific
failure. Faithfulness---citing real, on-signal features---is something the
baselines do nearly as well as we do: on P1 their faithfulness sits at
$2.23$--$2.31$ against our $2.69$. But the diagnosis those features are attached to
is wrong. P1 correctness separates the systems completely: $1.97$ for us against
$0.89$, $0.55$, and $0.00$ for Gemini, GPT-4o, and MedGemma. The general models are
fluent about the signal and confidently mistaken about the patient---a SUM would
let their faithfulness mask the failed diagnosis, and only the per-axis view makes
it visible. The same holds through P2 and P3, where using the metadata to place the
risk (context, our $2.03$ vs.\ Llama's $0.62$) and choosing a rhythm-specific
management (our $2.14$ vs.\ $1.29$--$1.85$) are what the baselines miss;
\textsc{Cardiologent} leads on every patient axis with no baseline topping any.
At the patient level, then, the baselines stop at a plausible-sounding account
while the judgment that a clinician would act on---the right diagnosis, tied to the
right patient context, with rhythm-specific management---is reached only by our
system; on the axis that matters most for a bedside decision, a confidently
written wrong answer is the baselines' typical failure, and the one our pipeline
is built to avoid.

\begin{table}[t]
\centering\small
\caption{\textbf{Patient-level diagnosis accuracy} ($n{=}65$): exact class, same
management family, and any mention. $\pm$ is a $95\%$ bootstrap CI ($B{=}10{,}000$).}
\label{tab:pdx}
\setlength{\tabcolsep}{5pt}
\begin{tabular}{lccc}
\toprule
& exact & family & mention \\
\midrule
\textbf{\textsc{Cardiologent}} & $\mathbf{0.48}{\scriptstyle\pm0.12}$ & $\mathbf{0.60}{\scriptstyle\pm0.12}$ & $\mathbf{0.77}{\scriptstyle\pm0.10}$ \\
GPT-4o         & $0.17{\scriptstyle\pm0.09}$ & $0.23{\scriptstyle\pm0.10}$ & $0.26{\scriptstyle\pm0.11}$ \\
Gemini-2.5-Pro & $0.12{\scriptstyle\pm0.08}$ & $0.15{\scriptstyle\pm0.08}$ & $0.34{\scriptstyle\pm0.12}$ \\
Qwen2.5-VL-72B & $0.08{\scriptstyle\pm0.07}$ & $0.12{\scriptstyle\pm0.08}$ & $0.25{\scriptstyle\pm0.11}$ \\
MedGemma-27B   & $0.00{\scriptstyle\pm0.00}$ & $0.00{\scriptstyle\pm0.00}$ & $0.22{\scriptstyle\pm0.10}$ \\
Llama-3.2-90B-Vision  & $0.00{\scriptstyle\pm0.00}$ & $0.00{\scriptstyle\pm0.00}$ & $0.00{\scriptstyle\pm0.00}$ \\
\bottomrule
\end{tabular}
\end{table}

\begin{table}[t]
\centering\footnotesize
\setlength{\tabcolsep}{3.2pt}
\caption{Patient-level score by the patient's dominant class: SUM over P1--P3,
max $24$. The \emph{all} row is the patient-level assessment summed over its three
tasks. Best per row in bold, second best underlined.}
\label{tab:patient-class}
\begin{tabular}{@{}lrcccccc@{}}
\toprule
class & $n$ & \textbf{\textsc{Cardiologent}} & gem. & gpt-4o & qwen & medg. & llama \\
\midrule
\multicolumn{8}{@{}l}{\emph{LLM judge}}\\
AF   & 13 & \textbf{21.4} & \underline{14.2} & 6.5 & 6.8 & 6.7 & 7.2 \\
SVT  & 13 & 6.8 & \textbf{7.9} & 6.2 & 6.9 & \underline{7.0} & 6.2 \\
VT   & 9  & 8.3 & \textbf{13.7} & \underline{11.7} & 6.1 & 5.4 & 6.0 \\
PVC  & 7  & \textbf{21.3} & 13.3 & \underline{14.3} & 10.0 & 7.7 & 5.4 \\
PAC  & 8  & \textbf{12.1} & 8.9 & \underline{10.4} & 7.4 & 7.4 & 7.1 \\
MAT  & 5  & \textbf{10.6} & \underline{8.0} & 7.0 & 7.6 & 6.8 & 7.0 \\
SND  & 7  & \textbf{15.9} & 6.7 & 7.0 & 7.1 & \underline{7.7} & 6.7 \\
AVB  & 3  & \textbf{10.7} & 6.0 & 6.0 & 6.3 & \underline{6.7} & 4.7 \\
all  & 65 & \textbf{13.6} & \underline{10.4} & 8.6 & 7.2 & 6.9 & 6.5 \\
\midrule
\multicolumn{8}{@{}l}{\emph{Cardiologists} ($20$-patient subset)}\\
AF   & 4 & \textbf{20.4} & \underline{19.4} & 6.1 & 6.1 & 6.5 & 4.8 \\
SVT  & 4 & \textbf{13.5} & \underline{10.0} & 5.8 & 4.8 & 7.9 & 4.9 \\
VT   & 3 & 6.0 & \underline{14.5} & \textbf{16.3} & 5.7 & 7.3 & 6.0 \\
PVC  & 2 & \underline{22.0} & 12.0 & \textbf{22.5} & 14.0 & 7.5 & 5.2 \\
PAC  & 2 & \underline{12.8} & 10.0 & \textbf{17.2} & 11.8 & 4.8 & 5.0 \\
MAT  & 2 & \textbf{10.2} & 4.0 & \underline{6.0} & 5.8 & 4.2 & 4.8 \\
SND  & 2 & \textbf{12.8} & 3.5 & 3.8 & 5.0 & \underline{9.5} & 4.0 \\
AVB  & 1 & \textbf{17.0} & 6.0 & \underline{10.5} & 4.0 & 4.5 & 8.0 \\
all  & 20 & \textbf{14.3} & \underline{11.3} & 10.3 & 6.9 & 6.8 & 5.1 \\
\bottomrule
\end{tabular}
\end{table}

\begin{table}[t]
\centering\small
\caption{Rater agreement per task, ICC(2,1) with absolute agreement ($\pm$ its
$95\%$ CI), on the $20$-patient subset. R1, R2 are the two cardiologists.}
\label{tab:icc-task}
\begin{tabular}{@{}lccc@{}}
\toprule
& R1\,$\leftrightarrow$\,R2 & Judge\,$\leftrightarrow$\,R1 & Judge\,$\leftrightarrow$\,R2 \\
\midrule
P1 (diagnosis)    & $0.711{\scriptstyle\pm0.105}$ & $0.651{\scriptstyle\pm0.117}$ & $0.669{\scriptstyle\pm0.140}$ \\
P2 (significance) & $0.612{\scriptstyle\pm0.146}$ & $0.714{\scriptstyle\pm0.127}$ & $0.549{\scriptstyle\pm0.153}$ \\
P3 (management)   & $0.480{\scriptstyle\pm0.170}$ & $0.518{\scriptstyle\pm0.172}$ & $0.488{\scriptstyle\pm0.166}$ \\
\bottomrule
\end{tabular}
\end{table}

\begin{figure}[!ht]
\centering
\includegraphics[width=\linewidth]{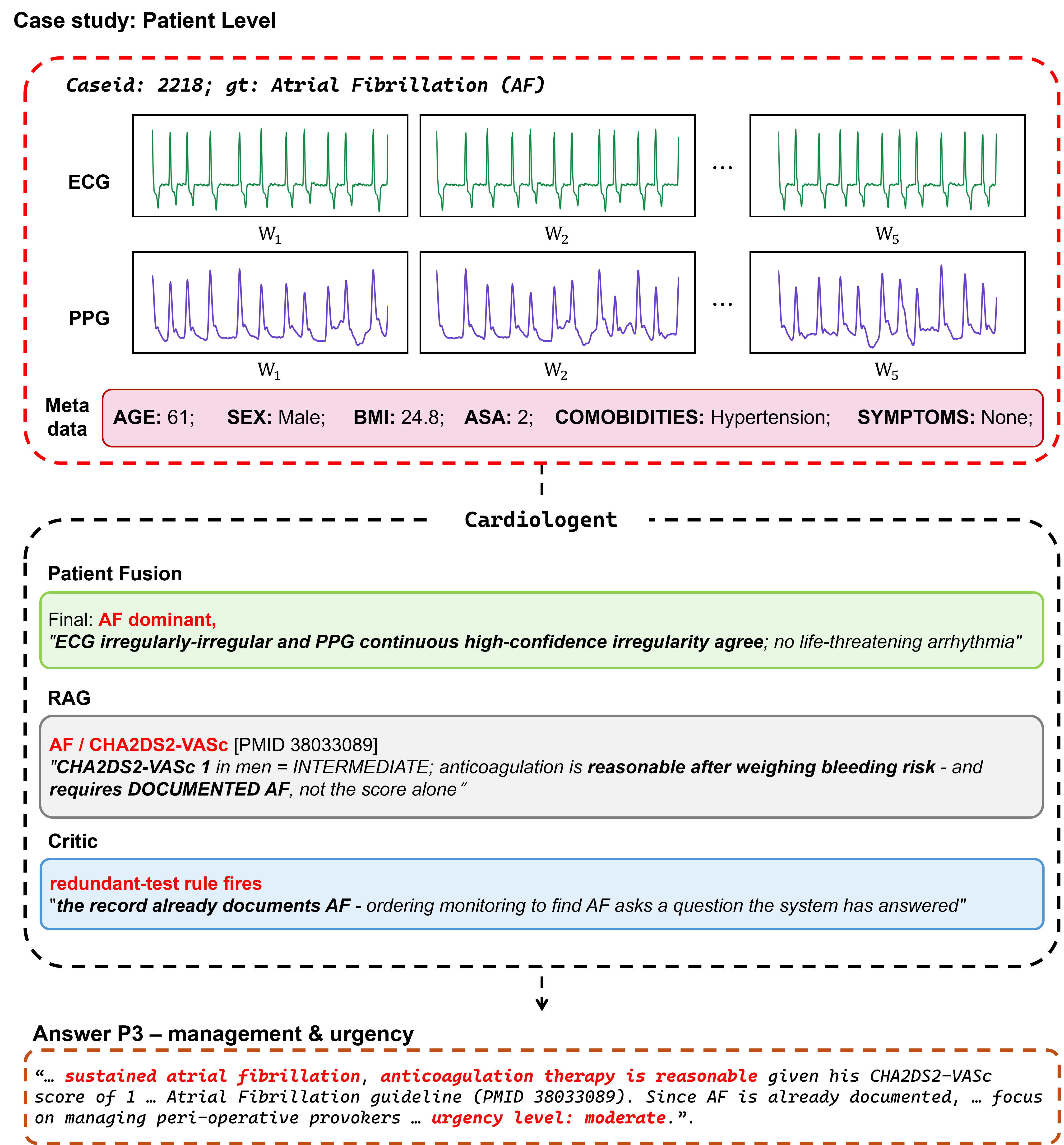}
\caption{\textbf{Case study: Patient-level} (case 2218; GT: atrial fibrillation).
The ECG and PPG readings agree, so patient fusion returns AF as the dominant
rhythm without a conflict to settle. Retrieval supplies the guideline that sets
how strongly the recommendation may be put---CHA$_2$DS$_2$-VASc~$1$ in a man is
intermediate, and anticoagulation is licensed by documented AF rather than by the
score---and the critic removes the draft's request for monitoring to find an AF
the record already documents. Excerpts are verbatim; ellipses mark omitted text.}
\label{fig:casestudy-patient}
\end{figure}

\paragraph{Patient-level diagnosis accuracy.}
Table~\ref{tab:pdx} scores the dominant rhythm at three grades: \emph{exact}, the
ground-truth dominant rhythm named (or a single-window dangerous finding
correctly flagged as suspected); \emph{family}, the named rhythm sharing a
management tier with the truth; and \emph{mention}, the true arrhythmia named
anywhere in P1. The families group the rhythms that share a management tier:
\{VT,\,MAT\}, \{PVC,\,PAC\}, and \{SND,\,AVB\}, with AF, SVT, and normal standing
alone, following the ventricular, supraventricular, and bradycardia management
guidelines~\citep{alkhatib2018va,page2016svt,kusumoto2019brady}. Unlike the
rubric scores, this is a hard label match, so it checks the P1 result against
something a graded judgement cannot inflate.
 
\textsc{Cardiologent} names the exact dominant rhythm for $0.48$ of patients
against $0.17$ for the strongest baseline, and the ordering is the same at every
grade. More telling is how the three grades move. Relaxing \emph{exact} to
\emph{family} gains us $0.12$ ($0.48{\to}0.60$), so a good part of our residual
error stays inside the management tier and would not change what is done; for the
baselines the same relaxation gains almost nothing (GPT-4o $0.17{\to}0.23$,
Gemini $0.12{\to}0.15$), meaning that when they are wrong they are wrong outside
the tier. Relaxing again to \emph{mention} separates two failure modes: Gemini
($0.15{\to}0.34$) and MedGemma ($0.00{\to}0.22$) do often name the true rhythm
somewhere in P1 but not as the dominant one---the finding is present and
mis-ranked---whereas Llama never names it at all. Our own gap between
\emph{exact} and \emph{mention} ($0.48$ vs.\ $0.77$) is the same failure in
milder form: for roughly a third of patients the true arrhythmia appears in the
impression without being made dominant.
 
\paragraph{Patient-level score by class.}
Table~\ref{tab:patient-class} breaks the patient-level total (P1--P3, max $24$)
out by each patient's dominant rhythm. \textsc{Cardiologent} leads six of the
eight classes, by the widest margins where the window level reads the rhythm
well: AF $21.4$ against $14.2$, PVC $21.3$ against $14.3$, SND $15.9$ against
$7.7$. The two it does not lead are VT ($8.3$ against Gemini's $13.7$) and SVT
($6.8$ against $7.9$)---the two rhythms the window level detects least reliably
($0.00$ and $0.08$ F1 on \emph{both}). A patient-level decision inherits the
window readings it is assembled from, and where those fail the decision does too.
Per-class counts are small ($3$--$13$ patients), so the ordering carries more than
the individual values. The cardiologist block, on the $20$-patient subset, follows
the same pattern---\textsc{Cardiologent} first overall and on most classes, VT the
clearest loss under both raters---with the counts too small ($1$--$4$) to read
individual classes closely.
 
\paragraph{Rater agreement by task.}
Table~\ref{tab:icc-task} breaks the overall judge--cardiologist ICC(2,1) out by task.
Agreement falls from P1 to P3 for the judge and cardiologists alike---the
integrated diagnosis is scored most consistently, and management least, where
even the two cardiologists agree only moderately ($0.48$). Management is where
clinical judgement legitimately diverges---the same rhythm admits more than one
defensible plan---so the lower agreement there reflects the task rather than the
judge: where the judge is noisier is exactly where the two cardiologists disagree
more as well. \textsc{Cardiologent} is still scored highest on P3 by both raters,
so its lead holds even on the task clinicians agree on least. At every task the
judge--cardiologist agreement overlaps the human--human agreement within its
confidence interval.

\paragraph{Case study: window fusion.}
Figure~\ref{fig:casestudy} traces a single window (GT: atrial fibrillation) through
the fusion, and shows why the two-signal read is more than a vote. The ECG agent
calls MAT---it sees irregular R--R and a narrow QRS but reads the P~wave as
present---while the PPG agent, which cannot be misled by an ambiguous P~wave, calls
AF from the irregularity alone ($\max|\Delta RR|\,0.47$\,s). The two conflict, so
fusion opens a debate rather than averaging. In the first round all three analyst
lenses (rate/rhythm, atrial activity, morphology) favor AF: no ventricular ectopy,
narrow QRS, and marked irregularity. The devil's advocate raises the one real
objection---an apparent P-ratio could mean organized atrial activity, and the
PPG cannot see P~waves, so its AF is the less anatomically grounded of the two
claims. In the second round the analysts weigh that against the measured
irregularity (R--R CV $0.274$/$0.234$) and hold AF, and the referee returns AF.
The correct label here is not the more confident agent's by default; it survives
because the objection was voiced and then outweighed by evidence, which is the
property the per-sample grounding and the per-axis reasoning scores reward
throughout this section.

\paragraph{Case study: patient level.}
Figure~\ref{fig:casestudy-patient} traces one record (case 2218; GT: atrial
fibrillation) through the three patient-level stages. \emph{Patient fusion} has
no conflict to settle: the ECG is irregularly irregular and the PPG reports
continuous irregularity with high confidence, so the two agree and the profile
comes back as AF dominant with no life-threatening rhythm---the work falls to the
stages after it. \emph{Retrieval} then supplies not the diagnosis but the
strength at which a recommendation may be put: for a man with a
CHA$_2$DS$_2$-VASc of $1$ the score is intermediate, so anticoagulation is
\emph{reasonable} rather than indicated, and it is licensed by documented AF
rather than by the score alone. The \emph{critic} deletes a step the draft had
added---an ambulatory monitor to screen for occult AF---because the record
already documents AF, and ordering a test to settle a question the system has
answered only defers the decision that question was meant to inform. What reaches
P3 is the therapy at that strength, without the confirmatory monitoring, at
moderate urgency.

\end{document}